\documentclass[letterpaper, 10 pt, conference]{ieeeconf}  % Comment this line out if you need a4paper

\IEEEoverridecommandlockouts                              % This command is only needed if 
\usepackage{graphics} % for pdf, bitmapped graphics files
\usepackage{epsfig} % for postscript graphics files
\usepackage{mathptmx} % assumes new font selection scheme installed
\usepackage{times} % assumes new font selection scheme installed
\usepackage{amsmath} % assumes amsmath package installed
\usepackage{amssymb}  % assumes amsmath package installed
\usepackage{siunitx} % for SI units
\usepackage{textcomp}
\usepackage{gensymb} % for degree command
\usepackage[colorlinks=true]{hyperref}
\usepackage{booktabs}
\usepackage{multirow}
\usepackage{xcolor}
\usepackage{booktabs}
\let\labelindent\relax
\usepackage{enumitem}
\usepackage{xspace}
\usepackage{booktabs}
\usepackage{soul}
\usepackage[noadjust]{cite}
\usepackage{makecell}
\usepackage{multicol}
\usepackage{rotating}
\usepackage{subcaption}
\usepackage[font=small]{caption}
\usepackage[percent]{overpic}

\usepackage{microtype}
\usepackage{contour}
\usepackage{courier}

\usepackage{enumitem}

\makeatletter
\DeclareRobustCommand\onedot{\futurelet\@let@token\@onedot}
\def\@onedot{\ifx\@let@token.\else.\null\fi\xspace}

\def\etal{et al\onedot}

\makeatother

\definecolor{MyDarkBlue}{rgb}{0,0.08,1}
\definecolor{MyDarkGreen}{rgb}{0.02,0.6,0.02}
\definecolor{MyDarkRed}{rgb}{0.8,0.02,0.02}
\definecolor{MyDarkOrange}{rgb}{0.40,0.2,0.02}
\definecolor{MyPurple}{RGB}{111,0,255}
\definecolor{MyRed}{rgb}{1.0,0.0,0.0}
\definecolor{MyGold}{rgb}{0.75,0.6,0.12}
\definecolor{MyDarkgray}{rgb}{0.66, 0.66, 0.66}

\newcommand\update[1]{{#1}}
\newcommand\finalupdate[1]{{#1}}
\newcommand{\mypara}[1]{\par\vspace*{0mm} \textbf{{#1}}}

\def\OURS{UMPNet\xspace}
\def\OURSSingleStep{SingleStep\xspace}
\def\OURSSgnOnly{AoTOnly\xspace}
\def\OURSMagOnly{SignedDist\xspace}
\def\OURSHP{UMPNet+HP\xspace}
\def\INVERSE{Inverse\xspace}
\def\WHERETOACT{Where2Act\xspace}
\def\WHERETOACTHP{Where2Act+HP\xspace}

\title{\LARGE \bf \OURS: Universal Manipulation Policy Network \\ for Articulated Objects \vspace{-2mm}}

\author{
  Zhenjia Xu \quad Zhanpeng He \quad Shuran Song \thanks{This work was supported by National Science Foundation under CMMI-2037101 and Amazon Research Award. We would like to thank Google for the UR5 robot hardware. Any opinions, findings, and conclusions or recommendations expressed in this material are those of the authors and do not necessarily reflect the views of the National Science Foundation.}\\
  Columbia University\\
 \href{https://ump-net.cs.columbia.edu/}{https://ump-net.cs.columbia.edu/}
}

\begin{document}

\maketitle
\thispagestyle{empty}
\pagestyle{empty}

\begin{abstract}
We introduce the Universal Manipulation Policy Network (UMPNet) -- a single image-based policy network that infers closed-loop action sequences for manipulating articulated objects. To infer a wide range of action trajectories, the policy supports 6DoF action representation and varying trajectory length. 
To handle a diverse set of objects, the policy learns from objects with different articulation structures and generalizes to unseen objects or categories.    
The policy is trained with self-guided exploration without any human demonstrations, scripted policy, or pre-defined goal conditions.  
To support effective multi-step interaction,  we introduce a novel Arrow-of-Time action attribute that indicates whether an action will change the object state back to the past or forward into the future. With the Arrow-of-Time inference at each interaction step, the learned policy is able to select actions that consistently lead towards or away from a given state, thereby, enabling both effective state exploration and goal-conditioned manipulation.

\end{abstract}
% \begin{IEEEkeywords}
% Deep Learning in Grasping and Manipulation, Perception for Grasping and Manipulation
% \end{IEEEkeywords}

\section{Introduction}

\begin{figure}
  \includegraphics[width=0.99\linewidth]{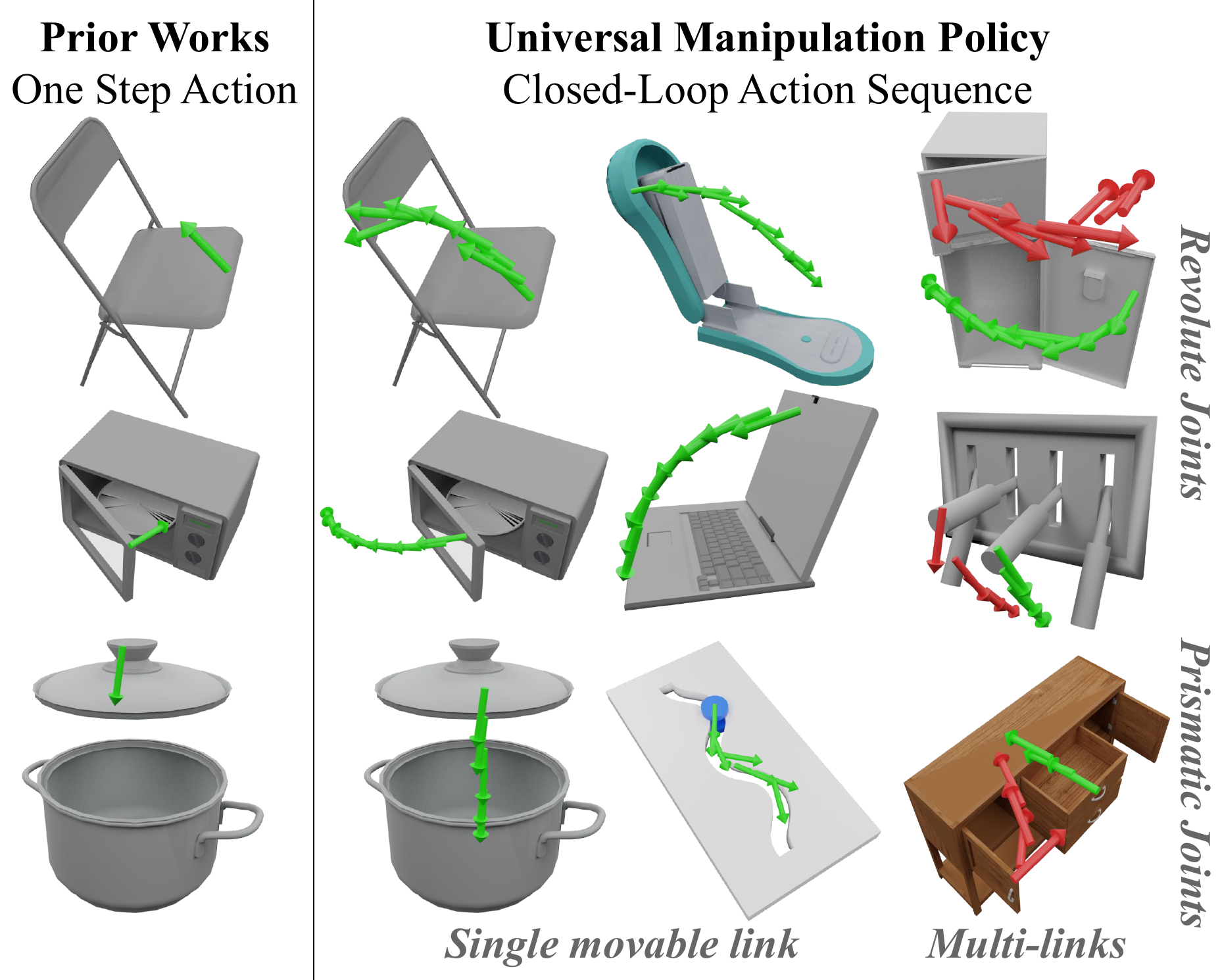}
    \vspace{-1mm}
    \caption{\textbf{Universal Manipulation Policy  for Articulated Objects.}
    Instead of predicting a single step action, \OURS predicts complex closed-loop 6DoF action sequences with varying trajectory length. As a result, the same policy network is able to handle a diverse set of objects regardless their joint types or number of links.}
    \label{fig:teaser}
    %https://docs.google.com/drawings/d/1b8YxmQLWy9RgRrzUp391ugzlIsbXLAyh1BrMUT1TeN0/edit
    \vspace{-7mm}
\end{figure}

The ability to effectively interact and manipulate unknown articulated objects is critical for many robotics tasks. 
However, due to the large variance in the objects' kinematic structure and 3D geometry, the actual action trajectories can vary drastically across different object instances and categories. 
Fig. \ref{fig:teaser} shows examples of action trajectories conditioned on different objects for opening a door, turning a switch, or opening a drawer. Extensive prior works have studied how to manually design or learn an object-specific policy for each type of interaction (e.g., opening doors). However, such policies are often time-consuming to design and fail to generalize across objects with different articulation structures.

While these interaction sequences are drastically different in their low-level geometric trajectories, many of them can be summarized by a similar high-level function conditioned on the objects' underlying geometric and kinematic structure.  
For example, the motion trajectory of a door opening can be represented by a function conditioned on its frame size and its rotation axis, and a similar function can also be used for opening a fridge, a microwave, or even a laptop. By learning to interact with a diverse set of articulated objects, the system is able to acquire a generalizable knowledge about objects' articulation structure and how these structures would react to different actions. Such knowledge goes beyond a specific object instance or category, allowing a universal interaction policy for any articulated objects.  

%On the other hand, the trajectory of opening a drawer will need another set of parameters due to its different joint properties. %Therefore, manually designing a policy for each type of interaction with each type of object would be extremely time-consuming.

Can we enable a robot to automatically acquire these basic concepts about the object structure through self-supervised interactions and use them to infer the corresponding manipulation policies? 
In this paper, we introduce the \textbf{Universal Manipulation Policy Network (\OURS)} -- a single policy network that discovers possible manipulation policies for an articulated object from visual observations (i.e., RGB-D images). The action trajectories inferred by the policy network (shown in Fig. \ref{fig:teaser}) highlight the following attributes: 
\vspace{-1mm}
\begin{itemize}[leftmargin=*]
    \item \ul{General action representation:} In order to model \textit{all} possible actions for \textit{any} articulated object, the network should be able to represent a general action space with little constraints -- it should be able to represent continuous actions in SE(3) with arbitrary trajectory length. 
    To achieve this goal, we formulate an action trajectory by its initial 3D position and a sequence of action directions, which allows the network to describe complex motion trajectories with varying sequence lengths. 
    % how to parameters a general motion primitive that can describe then 
    
    \item \ul{Closed-loop action sequence:}  Instead of predicting a single step action (e.g., push or pull), we are interested in predicting long-horizon sequential actions that could describe a complex motion trajectory. However, due to error accumulation and partial observation, directly predicting the full trajectory from the initial state can be challenging. %especially when the object's visual observation can be highly incomplete due to self-occlusion.  
    To address this issue, we use a closed-loop formulation where the network continues to predict the next action conditioned on the object's initial and current state, allowing the network to adjust its action prediction based on its visual observation of the object. 
    
    \item \ul{Arrow-of-Time awareness:} Most of the action trajectories are bi-directional in time (i.e., they are valid in either direction). Hence, conditioning on a single state can result in multiple effective next actions that would change the object's state with the same magnitude. 
    However, to avoid the back-and-forth actions, the network takes the history state as input and infers an additional ``Arrow-of-Time (AoT)'' attribute for each action. This AoT label indicates whether this action will change the object state back to the past or forward into the future.   Apart from encouraging exploring new states, this Arrow-of-Time inference also allows us to \textit{directly} apply the network in ``goal conditioned manipulation'', where we can simply swap out the initial state with the goal state and choose the actions using a reversed Arrow-of-Time. 
    
\end{itemize}

\update{In summary, we present a unified framework that discovers possible manipulation policies for  an articulated object from visual observations.}  By using self-guided exploration, the policy network is able to learn a wide range of action trajectories for a diverse set of objects and generalize to unseen objects and categories. 
The training does not require any human demonstrations or pre-defined goal conditions.
We validate our approach on two manipulation tasks (1) open-ended state exploration and (2) goal-conditioned manipulation. The experiments demonstrate that  \OURS is able to outperform alternative approaches in both tasks significantly. %Finally, we can also extract the policy's implicit belief about the objects articulation structure by computing the estimated joint parameter (e.g., joint axis and potion) from the  actions selected by the trained policy. 

\section{Related work}

\mypara{Open-loop manipulation with pose estimation.}
Many works have focused on learning task-specific manipulation primitives, such as grasping \cite{song2020grasping}, pushing \cite{li2018push} and tossing \cite{zeng2019tossingbot}.
For articulated objects, methods have focused on handling doors, and drawers \cite{klingbeil2010learning, abbatematteo2019learning, jainscrewnet20, mittal2021articulated, jain2020learning, ruhr2012generalized, harada2019service, schmid2008opening, kessens2010utilizing}.
These prior works typically start with object pose estimation \cite{gadre2021act, li2019articulated-pose} and then use the object pose to compute an open-loop motion trajectory.   However, the action trajectory designed for one task (e.g., opening doors) may be too specific to be applied to other objects or tasks (e.g., pushing button). Moreover, performing pose estimation for articulated objects with \textit{unknown category and kinematic structure} is an extremely challenging task.  
On the contrary, our model does not require any object detection, pose estimation or part segmentation, and demonstrates that it is in fact not \textit{necessary} to perform explicit pose estimation to perform effective manipulations. 
%, since the objects’ underlying geometry and structure would already inform the manipulation policy
%universally models all possible action trajectories to handle different kinds of articulated objects. 
%, ruhr2012generalized, harada2019service, schmid2008opening, kessens2010utilizing

\mypara{Learning action trajectories from demonstrations.}
Another popular method for robots to acquire new manipulation skills is learning from demonstrations. %\cite{agrawal2016learning}.
This approach has been explored extensively in reinforcement learning literature \cite{niekum2013incremental}.
Researchers has tried using behavioral cloning to learn from human demonstration data captured by various methods, for example, motion capture \cite{kober2010imitate, peter2009generalization}, videos \cite{sermanet2017contrastive, huang2018neural, huang2020motion, dang2010robot} and virtual reality \cite{zhang2017imitate, lynch2020learning}.
\update{However, these works requires collection of large amount of high quality demonstrations with action and pose annotation, which is expensive to obtain.}
In contrast, our framework generates its own training data by allowing the agent to actively interact with objects and explore the environment.

\mypara{Single-step action affordance.}
Action affordance describes the possibility of an action to be applied to a given location in the environment. The task of affordance prediction does not limit to a specific kind of object or action primitive.  Building on the well-studied image segmentation problems,  many existing methods have been developed to learn object affordance through passive observations, such as learning human-object interaction hotspots from video \cite{interaction-hotspots, nagarajan2020ego-topo} and contact heatmap from RGB-D image \cite{brahmbhatt2019contactdb}. % Apart from of low-level atomic actions, Nagarajan and Grauman \cite{interaction-exploration} propose to learn high-level affordance concepts (e.g., sliceable). 
The work most related to ours is ``Where2Act'' by Mo \etal \cite{mo2021where2act}, where the algorithm can infer single step action affordance for different articulated objects. 
However, limited by its single step formulation, this approach fails to generated long-horizon motion trajectories for goal-conditioned manipulation tasks, which is the focus of our approach.

\section{Approach}
\begin{figure}
    \vspace{2mm}
    \centering
    \includegraphics[width=\linewidth]{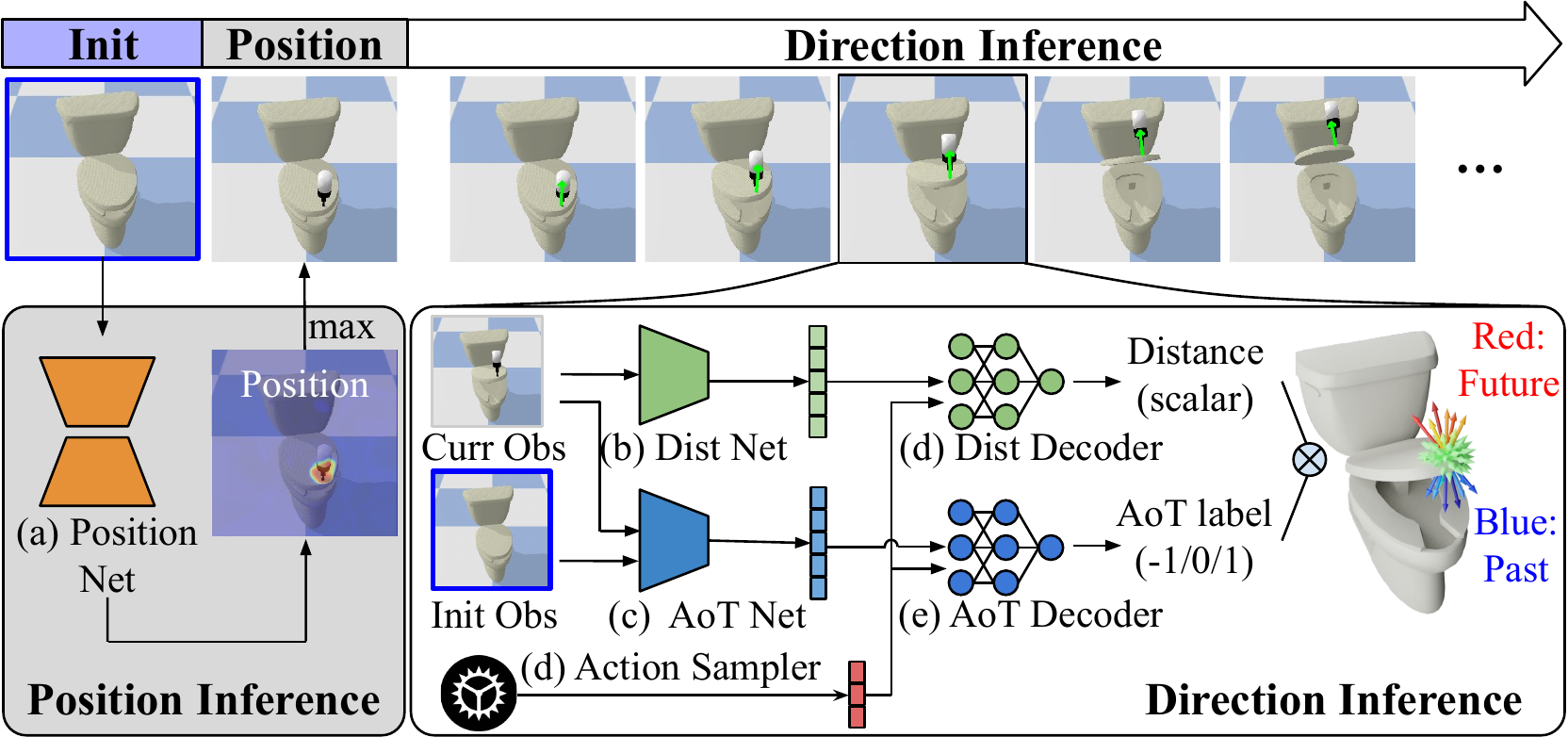}
    \caption{\textbf{Approach overview.} \OURS takes visual observation (i.e., RGB-D images) of an articulated object as input and generates a sequence of actions in SE(3) space to explore novel object states. \textbf{(Left)} A grasp position is selected in the first interaction step.   \textbf{(Right)} In following steps, the outcomes for each action candidates ($r_\textrm{dist}$ and $r_\textrm{AoT}$)  are inferred and then used for action direction selection.  $r_\textrm{dist}(a^\mathrm{dir}_t)$ infers the potential  moving distance of the joint after applying the action $a^\mathrm{dir}_t$. 
    $r_\textrm{AoT}(a^\mathrm{dir}_t)$ infers whether or not the action will move the object toward a novel state.  
    The action direction with largest $r_\textrm{dist}$ and positive $r_\textrm{AoT}$ will be selected. 
    }
    \label{fig:model}
    % https://docs.google.com/drawings/d/1V67vtjn-8yXD9R2tTJGalmurSRVxHjcX-piK0URPnGA/edit?usp=sharing
    \vspace{-6mm}
\end{figure}

The goal of the manipulation policy $\pi$ is to generate a sequence of actions to interact with a random articulated object which would result in novel states that haven't been visited before. 
Taking Fig. \ref{fig:model} as an example, to effectively explore novel states of the object (i.e., a toilet),  the algorithm should be able to (a) choose the right position on the object to interact with (i.e.,  interacting with the cover instead of the base), (b) select a proper action direction (i.e.,  pulling up instead of pushing down), and (c) consistently select actions in the following steps to explore novel states (i.e., keeping pulling up the cover instead of moving up-and-down). 
These three requirements directly correspond to the three key components of our algorithm, which are action position selection (a), action distance (b) and Arrow-of-Time inference (c) for action direction selection. 
As a result, the final system is able to learn through a self-guided exploration process, without explicit human demonstrations \cite{lynch2020learning}, scripted policy \cite{mo2021where2act}, or pre-defined goal conditions \cite{nasiriany2019planning}.

%
%The following sections provide a formal problem definition (\S \ref{sec:problem}),  details on policy network (\S \ref{sec:position}-\S \ref{sec:aot}) followed by training details (\S \ref{sec:training}) and how to apply the learned policy to goal-conditioned manipulation task (\S~\ref{sec:manipulation}).

\vspace{-1mm}
\subsection{Problem formulation}
\vspace{-0.5mm}
\label{sec:problem}
The task is defined as follows: given a visual observation of an articulated object in the form of an RGB-D image at the initial and current state $o_0, o_t \in \mathbb{R}^{W \times H \times 4}$, the agent with a policy $\pi$ generates an action $a_t$ at each step $\pi(o_t, o_0) \rightarrow a_t$ that satisfies the aforementioned requirements. 
The action is represented in SE(3) space, parameterized by end-effector (i.e., a suction-based gripper) position and moving direction $a_t = (a^\mathrm{pos}_t, a^\mathrm{dir}_t)$, where $a^\mathrm{pos}_t \in \mathbb{R}^3$ is a 3D coordinate and $a^\mathrm{dir}_t \in \mathbb{R}^3 , (||a^\mathrm{dir}_t||=1)$ is a unit vector in 3D indicating the end-effector moving direction.

In the first interaction step, the policy selects a 3D position $a^\mathrm{pos}_0$ to apply action (i.e., an immobilizing grasp via suction). To execute the action, the agent moves its end-effector to this position, with an orientation perpendicular to the object surface. \update{Note that the gripper orientation (determined by the surface normal) can be different from the action direction $a^\mathrm{dir}_t$ (determined by the Direction Inference Networks Sec. \ref{sec:direction})}. In each following step, the agent will select a 3D direction $a^\mathrm{dir}_t$ and move its end-effect 0.18(m) along that direction,  the position $a^\mathrm{pos}_t$  is fixed relative to the objects surface. The suction behavior is implemented as a force constraint between the suction cup and the selected 3D position on the object. The orientation of the end-effector is always aligned with the surface normal during the interaction.

\vspace{-1mm}
\subsection{Position inference}
\vspace{-0.5mm}
\label{sec:position}

To start, the policy needs to determine a suitable position on the object 3D surface  $a^\mathrm{pos}_0$  to apply action (i.e., a immobilizing grasp via suction). 
To do so, the algorithm needs to select a pixel from the observation image $o_0$ to apply action. The selected pixel will then be projected back to the 3D space using the depth value provided in the RGB-D image.  

We formulate this problem as an image labeling task, where the position network (Fig. \ref{fig:model}a) takes in an RGB-D image and predicts per-pixel position affordance score $P \in [0,1]^{W \times H}$. The affordance score $P(w, h)$ implies the likelihood of the object part movement when applying an action in this position. 
We use a U-Net architecture for this task, the network is supervised by the outcome of the executed action (one out of $W \times H$ pixels).  The ground truth label is $1$ if and only if the object state is changed in any of the future steps. The network is trained with Binary Cross-Entropy loss. 

Note that simply selecting a position belonging to a movable link is a necessary but not sufficient criteria.  
For example, if the selected position is very close to the joint axis, the agent will not be able to apply enough force to move the object part. 
Furthermore, the label is affected by the quality of direction selection. A correct position can still be labeled as a negative case if the object state is not changed due to wrong direction predictions in the following steps.

\vspace{-1mm}
\subsection{Direction inference}
\vspace{-0.5mm}
\label{sec:direction}

At this point, the end-effector has grasped the object link at $a^\mathrm{pos}_0$ which is visible to the camera. Conditioned on this information, the policy then needs to select a 3D direction $a^\mathrm{dir}_t$. 
\update{To select the action direction, the algorithm need to first sample a set of action candidates, and evaluate each action candidate's effectiveness. The ``effectiveness'' is measured by the moving distance of the object joint position $r_\mathrm{dist}(a^\mathrm{dir}_t)$ and Arrow-of-Time attribute $r_\mathrm{AoT}(a^\mathrm{dir}_t)$, defined as following:}
\[r_\mathrm{dist}(a^\mathrm{dir}_t) = ||\vec{j}_t - \vec{j}_{t - 1}|| \]
\[\gamma = (\vec{j}_t - \vec{j}_{t - 1}) \cdot (\vec{j}_{t-1} - \vec{j}_0) \]
\[r_\mathrm{AoT}(a^\mathrm{dir}_t) = 
\left\{
\begin{aligned}
\arraycolsep=1.2pt\def\arraystretch{1.3}
\begin{array}{rcl}
-1 &  & {r_\mathrm{dist}(a^\mathrm{dir}_t) > \update{\delta} ~\&~ \gamma < 0} \\
0 &  & {r_\mathrm{dist}(a^\mathrm{dir}_t) \leq \update{\delta}} \\
1 &  & {r_\mathrm{dist}(a^\mathrm{dir}_t) > \update{\delta} ~\&~ \gamma \geq 0}
\end{array}
\end{aligned}
\right.
\]
where $\vec{j}_t$ is the object joint state in each step $t$ and \update{$\delta$} is a threshold to determine whether the state is effectively changed. \update{$\delta$ is 0.15m for prismatic joint and 8.6\degree for revolute joint.}
% In each interaction step, we assume there is only one major active link that is interact by the agent.  
\update{The following paragraph provides details on how to generate action candidates $\{a_t^\mathrm{dir}\}$, and infer $r_\mathrm{dist}(a^\mathrm{dir}_t)$ and $r_\mathrm{AoT}(a^\mathrm{dir}_t)$}. 
%The goal of the policy is then to infer the outcome for all action candidates $\{\hat{a}^\mathrm{dir}\}$ generated by a direction sampler. The direction action $\hat{a}^\mathrm{dir}_t$ with the largest distance prediction ($\tilde{r}_\mathrm{dist}(\hat{a}^\mathrm{dir}_t)$) and positive AoT prediction ($\tilde r_\mathrm{AoT}(\hat{a}^\mathrm{dir}_t)$ ) will be selected. 

\update{To generate direction candidates $\{\hat{a}^\mathrm{dir}\}$}, one naive method would be uniformly sampling in the SO(3) space. 
However, limited by the number of samples, the sampled directions can only cover a small portion of the continuous action space that does not include the optimal directions.
To address this issue, we use a heuristic approach, iterative cross-entropy method (CEM), to reduce the sampling space to achieve efficient direction sampling. 
The algorithm starts with uniform sampling the SO(3) space for $N$ samples. Then, it evaluates the sampled actions based on the predicted action scores: $s(\hat{a}) = \tilde{r}_\mathrm{dist}(\hat{a}^\mathrm{dir}) \cdot \tilde{r}_\mathrm{AoT}(\hat{a}^\mathrm{dir})$. 
In the next iteration, the algorithm re-sampls the candidates with probability correlated to its score: $ p(\hat{a}) \propto e^{T * s(\hat{a})}$, where $T=20$ is a temperature value. Added a random noise, they are considered as candidates in the second interaction. In this way, the samples in the second iteration will concentrate on the region that has more "potential", leading to better performance with the same number of samples. Detailed comparisons are listed in appendix. Our final model uses CEM sampling with 64 samples.

%A naive method is uniform sampling on unit sphere. But the sparsity leads to poor sample efficiency because most of the actions won't make change object state effectively. To this end, a smarter way is cross-entropy method (CEM) with two stages. 
%The first stage is uniform sampling. In the second stage, we re-sample from the candidates in the first stage with probability correlated to its prediction: $Prob(a) \propto e^{T * s(a)}$, where 20 is chosen for the temperature $T$. Added a random noise, they are considered as candidates in the second stage. Therefore, more focus will be put on the region with more "potential" and the performance will be better with the same number of samples. Detailed comparison are listed in Sec. \ref{sec:exp:action_sample}. We employ CEM with 64 samples in all following experiments.

%\mypara{Distance inference.}
\update{To infer the moving distance $ \tilde{r}_\mathrm{dist}(a^\mathrm{dir}_t)$ for an action candidate, the network needs to consider the object's current state and grasp position which are both encoded in the current observation $o_t$.}
Taking in the RGB-D image of the current state, DistNet (Fig. \ref{fig:model}b) outputs embedding vector $\psi(o_t)$. Then DistDecoder (Fig. \ref{fig:model}d) takes both embedding vector $\psi(o_t)$ and action $a$ as input, and outputs a scalar as the distance prediction $ \tilde{r}_\mathrm{dist}(a^\mathrm{dir}_t)$. DistNet is a convolution neural network and the output is flattened to an embedding vector. Dist-Decoder is a fully-connected neural network trained using MSE loss $\mathcal{L}_{dist}$ for the executed action $a_t$. 

\begin{figure}[t]
    \vspace{0mm}
    \includegraphics[width=\linewidth]{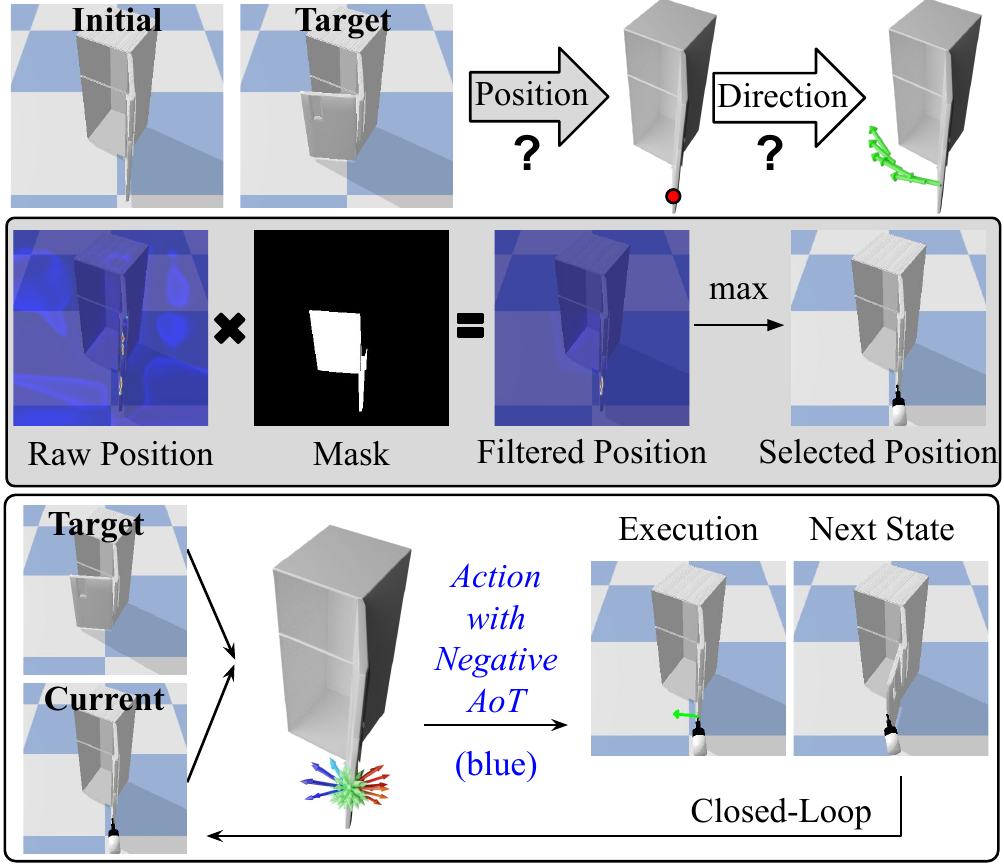}
    \caption{\textbf{Goal conditioned manipulation.}}
    \label{fig:goal-method}
    \vspace{-7mm}
\end{figure}

%\subsubsection{Arrow-of-Time inference.}
%\label{sec:aot}
% Different from distance inference, Arrow-of-Time Inference depends on both current observation and initial observation.
%Understanding the object structure and generating a single direction to move the object are not enough. Another major challenge is to avoid moving back-and-forth.
\update{
Different from  $ \tilde{r}_\mathrm{dist}(a^\mathrm{dir}_t)$ inference,
Arrow-of-Time $ \tilde{r}_\mathrm{AoT}(a^\mathrm{dir}_t)$ inference is conditioned on on both current observation and initial observation.}
For single-step interaction, any action that changes the object's state would result in a novel state. However, it is not true for multi-step interactions -- the policy can move the object link back-and-forth without exploring any new states. 
To address this issue, we proposes an ``Arrow-of-Time'' (AoT) action attribute that indicates whether the action will change the object state back to the initial state or forward into the future (i.e., a novel state). 
Specifically, AoTNet (Fig. \ref{fig:model}c) takes the current and initial observation as input and outputs another embedding vector $\phi(o_t, o_0)$. This embedding vector is then combined with the action embedding to infer the final AoT label for this action $ \tilde{r}_\mathrm{AoT}(a^\mathrm{dir}_t)$.
The network architectures of the AoT branch is similar to those of the Dist branch while the only differences are the different input dimensions of the Dist Net and the AoT Net as well as the different output dimensions of the AoT Decoder and the Dist Decoder. The model is trained as a three-way classification with Cross-Entropy loss $\mathcal{L}_{AoT}$.
The final loss for direction inference is: $\mathcal{L} = \lambda \mathcal{L}_{dist} + \mathcal{L}_{AoT}$, where  $\lambda = 100$ in our experiments.
\vspace{-1.5mm}
\subsection{Training}
\vspace{-0.5mm} 
\label{sec:training}
All training data come from interaction trials executed by the policy trained from scratch. A FIFO replay buffer (size=$6400$) is used to store training data.
To collect data with both positive and negative AoT labels, we employ contradictory policy for direction inference within a sequence. In the first half of each sequence, we select action with positive AoT prediction for execution to move the object away from its initial state. In the second half, actions with negative AoT prediction are executed to encourage the object to move back. 16 trajectories are collected in each epoch. The sequence length is 4 at the beginning. After 1000 epochs, it increases by 2 every 400 epochs, until reaching 20. $\epsilon$-greedy is used during training, where $\epsilon$ decreases linearly from $1$ to $\epsilon_{min}$ within $n$ epochs. In position inference, $n=300$ and $\epsilon_{min}=0.1$. In direction inference, $n=500$ and $\epsilon_{min}=0.2$. 

Position module and direction module are trained with 8 iterations accordingly in each epoch.
In each position training iteration, we sample a batch (size=$16$) of examples from the replay buffer with a 1:1 positive to negative ratio. 
In each direction training iteration, 1:1:1 samples from positive, negative, and not-moving data form a batch (size=$24$). 
%Optimization is carried out using ADAM \cite{Kingma2015Adam}. The initial learning rate is 8e-3 and will decay with a ratio of 0.5 every 500 epochs. The model is trained for 5000 epochs using a NVIDIA 3090 GPU.

\begin{table*}[t]
% \vspace{1.5mm}
\caption[]{Effective state exploration.\protect\footnotemark \label{tab:exploration}}
{\scriptsize 
    \centering
    \setlength\tabcolsep{2pt}
    \begin{tabular}{r|cccccccccccc|cccccccccc}
    \toprule
     & \multicolumn{12}{c|}{Novel instances in training categories } &  \multicolumn{10}{c}{Testing categories} \vspace{0.5mm}  \\ 
     %\hline
      & 
    \includegraphics[width = 0.032\linewidth]{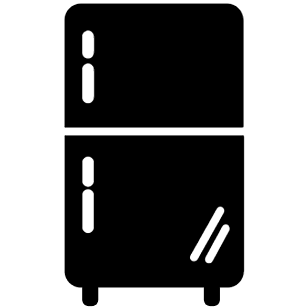} &
    \includegraphics[width = 0.032\linewidth]{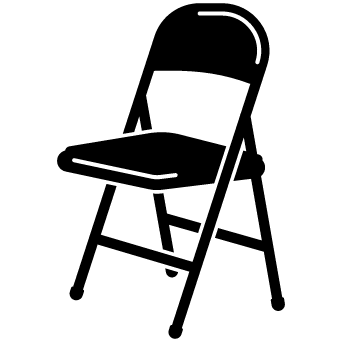} & 
    \includegraphics[width = 0.032\linewidth]{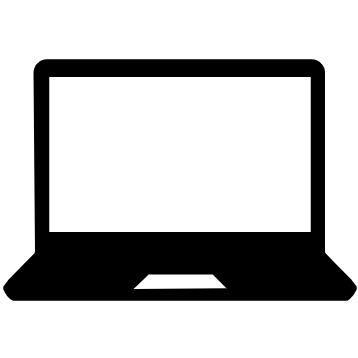} &
    \includegraphics[width = 0.032\linewidth]{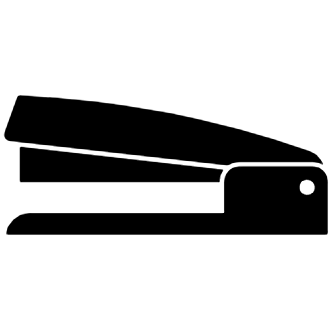} &
    \includegraphics[width = 0.032\linewidth]{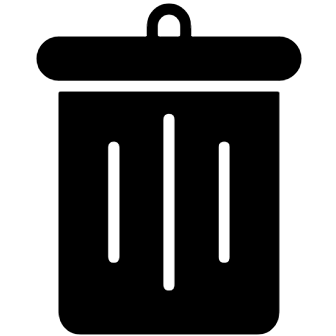} &
    \includegraphics[width = 0.032\linewidth]{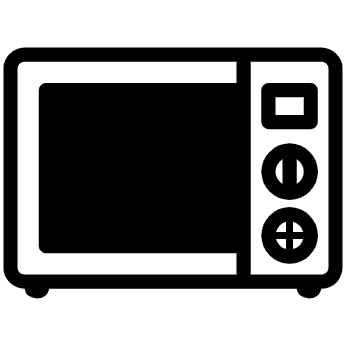} &
    \includegraphics[width = 0.032\linewidth]{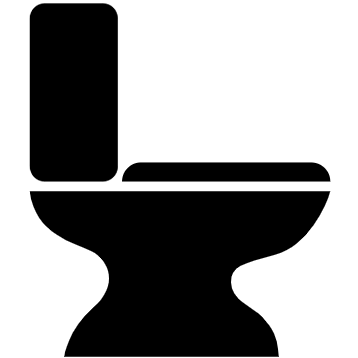} &
    \includegraphics[width = 0.032\linewidth]{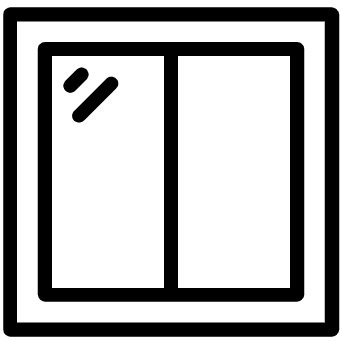} &
    \includegraphics[width = 0.032\linewidth]{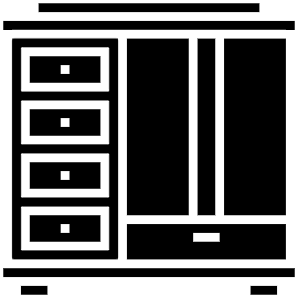} &
    \includegraphics[width = 0.032\linewidth]{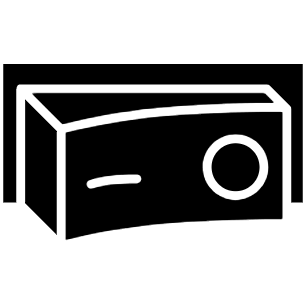} &
    \includegraphics[width = 0.032\linewidth]{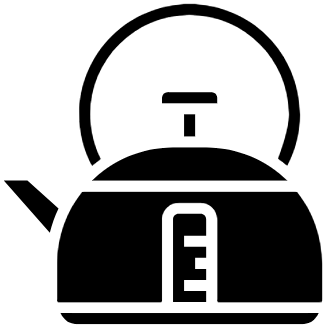} &
    \includegraphics[width = 0.032\linewidth]{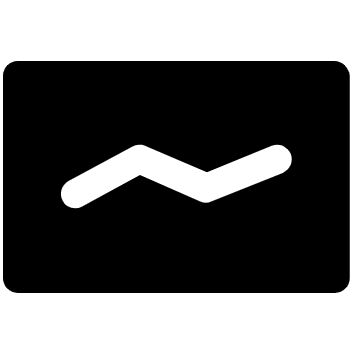} &
    \includegraphics[width = 0.032\linewidth]{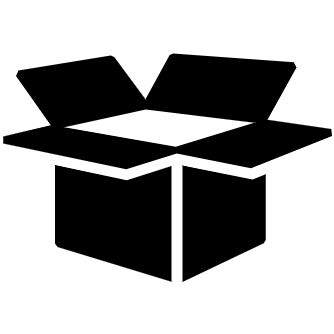} &
    \includegraphics[width = 0.032\linewidth]{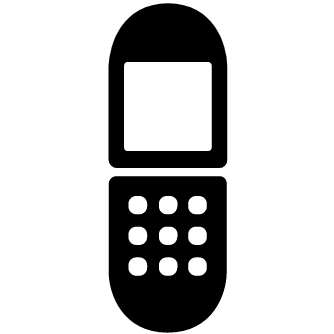} &
    \includegraphics[width = 0.032\linewidth]{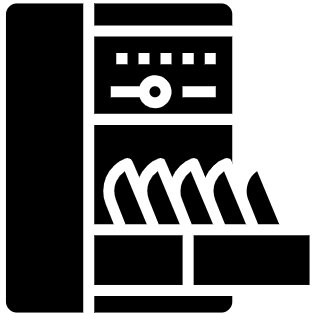} &
    \includegraphics[width = 0.032\linewidth]{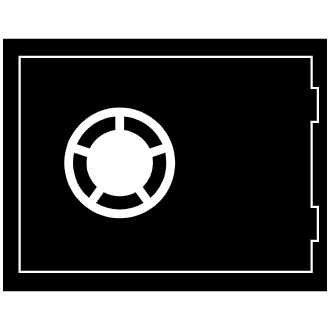} &
    \includegraphics[width = 0.032\linewidth]{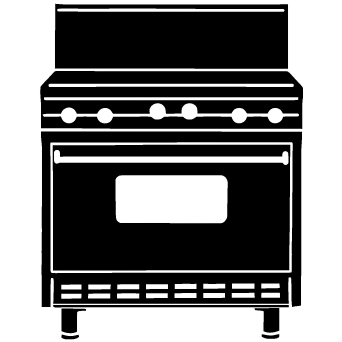} &
    \includegraphics[width = 0.032\linewidth]{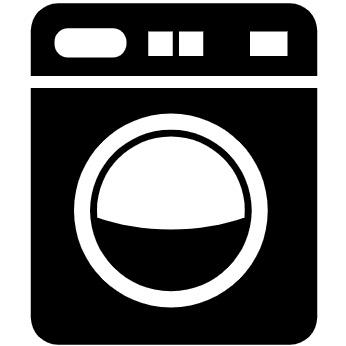} &
    \includegraphics[width = 0.032\linewidth]{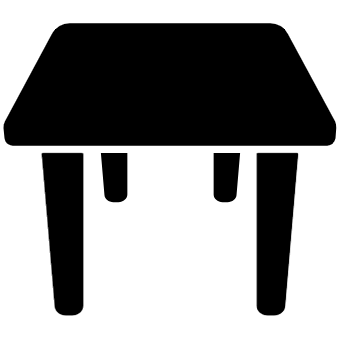} &
    \includegraphics[width = 0.032\linewidth]{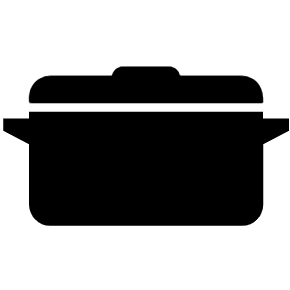} &
    \includegraphics[width = 0.032\linewidth]{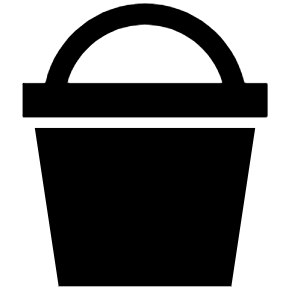} &
    \includegraphics[width = 0.032\linewidth]{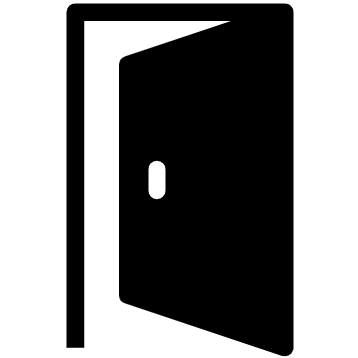}
    \vspace{0.5mm} \\
    
    %\midrule
     \WHERETOACT & 0.94 & \textbf{2.08} & 1.10 & \textbf{0.79} & \textbf{0.92} & 1.24 & 1.05 & \textbf{1.06} & 0.80 & \textbf{0.74} & \textbf{0.96} & 0.57 & 0.96 & 1.48 & 1.01 & 1.17 & 1.17 & \textbf{1.95} & \textbf{0.82} & 1.02 & 1.38 & 0.81 \\
    
     \OURSSgnOnly & 0.99 & 1.42 & 1.05 & 0.63 & 0.62 & 1.01 & 0.76 & 0.62 & 0.61 & 0.54 & 0.57 & 0.51 & 0.75 & 1.10 & 1.06 & 1.10 & 1.14 & 1.46 & 0.49 & 0.86 & 1.21 & 0.80 \\
     
     \OURSMagOnly & 0.84 & 1.68 & 1.04 & 0.53 & 0.91 & 1.25 & 1.23 & 0.69 & 0.73 & 0.43 & 0.65 & 0.51 & 0.75 & 1.10 & 1.06 & 1.10 & 1.14 & 1.46 & 0.49 & 0.86 & 1.21 & 0.80 \\
     
     \OURS & \textbf{1.02} & \textbf{2.08} & \textbf{1.37} & 0.73 & \textbf{0.92} & \textbf{1.29} & \textbf{1.26} & 1.03 & \textbf{0.81} & 0.70 & 0.90 & \textbf{0.66} & \textbf{1.10} & \textbf{1.50} & \textbf{1.14} & \textbf{1.18} & \textbf{1.32} & 1.87 & 0.77 & \textbf{1.05} & \textbf{1.69} & \textbf{0.90} \\
    \midrule

    \multicolumn{23}{c}{Single action effects $\uparrow$} \\
    
    \midrule
    \WHERETOACT & 0.38 & 0.45 & 0.34 & 0.25 & 0.52 & 0.56 & 0.49 & 0.56 & 0.45 & 0.50 & 0.58 & 0.26 & 0.39 & 0.39 & 0.45 & 0.42 & 0.51 & 0.53 & 0.50 & 0.66 & 0.24 & 0.34\\
    
    % \WHERETOACTHP & \textbf{0.72} & \textbf{0.85} & 0.89 & 0.48 & 0.60 & 0.83 & \textbf{0.85} & 0.72 & 0.62 & \textbf{0.63} & 0.73 & 0.50 & 0.75 & \textbf{0.87} & \textbf{0.79} & 0.84 & \textbf{0.81} & 0.89 & 0.54 & \textbf{0.86} & 0.91 & 0.65\\
    
    \WHERETOACTHP & \textbf{0.72} & 0.85 & 0.89 & 0.48 & 0.60 & 0.83 & \textbf{0.85} & 0.72 & 0.62 & \textbf{0.63} & 0.73 & 0.50 & 0.75 & 0.87 & \textbf{0.79} & 0.84 & 0.81 & 0.89 & 0.54 & 0.86 & 0.91 & 0.65\\
    
     \OURSSingleStep & 0.31 & 0.42 & 0.39 & 0.26 & 0.47 & 0.51 & 0.48 & 0.49 & 0.44 & 0.47 & 0.57 & 0.24 & 0.44 & 0.38 & 0.39 & 0.41 & 0.45 & 0.45 & 0.47 & 0.78 & 0.29 & 0.31\\

    \OURSSgnOnly & 0.58 & 0.77 & 0.69 & 0.42 & 0.47 & 0.68 & 0.62 & 0.67 & 0.50 & 0.44 & 0.59 & 0.44 & 0.70 & 0.76 & 0.65 & 0.82 & 0.61 & 0.81 & 0.44 & 0.80 & 0.83 & 0.50\\
     \OURSMagOnly & 0.43 & 0.59 & 0.66 & 0.38 & 0.47 & 0.54 & 0.58 & 0.58 & 0.46 & 0.38 & 0.48 & 0.38 & 0.60 & 0.57 & 0.51 & 0.58 & 0.57 & 0.65 & 0.36 & 0.55 & 0.68 & 0.47\\
    % \OURS & \textbf{0.70} & \textbf{0.85} & \textbf{0.90} & \textbf{0.52} & \textbf{0.60} & \textbf{0.87} & \textbf{0.81} & \textbf{0.74} & \textbf{0.64} & \textbf{0.55} & \textbf{0.74} & \textbf{0.52} & \textbf{0.77} & \textbf{0.85} & \textbf{0.76} & \textbf{0.85} & \textbf{0.80} & \textbf{0.92} & \textbf{0.56} & \textbf{0.86} & \textbf{0.93} & \textbf{0.68}\\
    
    % \OURS & 0.70 & \textbf{0.85} & \textbf{0.90} & \textbf{0.52} & \textbf{0.60} & \textbf{0.87} & 0.81 & \textbf{0.74} & \textbf{0.64} & 0.55 & \textbf{0.74} & \textbf{0.52} & \textbf{0.77} & 0.85 & 0.76 & \textbf{0.85} & 0.80 & \textbf{0.92} & \textbf{0.56} & \textbf{0.86} & \textbf{0.93} & \textbf{0.68}\\
    
    \OURS & 0.70 & 0.85 & 0.90 & 0.52 & 0.60 & 0.87 & 0.81 & 0.74 & 0.64 & 0.55 & 0.74 & 0.52 & 0.77 & 0.85 & 0.76 & 0.85 & 0.80 & 0.92 & 0.56 & 0.86 & 0.93 & 0.68\\
    
    \OURSHP & 0.71 & \textbf{0.86} & \textbf{0.90} & \textbf{0.57} & \textbf{0.64} & \textbf{0.88} & 0.83 & \textbf{0.74} & \textbf{0.65} & 0.60 & \textbf{0.74} & \textbf{0.55} & \textbf{0.77} & \textbf{0.88} & 0.78 & \textbf{0.86} & \textbf{0.83} & \textbf{0.92} & \textbf{0.56} & \textbf{0.88} & \textbf{0.93} & \textbf{0.70}\\
    
    \midrule
    
    % novel state number
    % \midrule
    % \WHERETOACT & 0.14 & 0.20 & 0.20 & 0.17 & 0.20 & 0.13 & 0.34 & 0.08 & 0.18 & 0.08 & 0.15 & 0.13 & 0.20 & 0.22 & 0.27 & 0.15 & 0.14 & 0.32 & 0.11 & 0.18 & 0.16 & 0.20\\
    %  \OURSSingleStep & 0.12 & 0.46 & 0.28 & 0.13 & 0.10 & 0.14 & 0.17 & 0.20 & 0.15 & 0.10 & 0.07 & 0.09 & 0.21 & 0.30 & 0.27 & 0.27 & 0.20 & 0.52 & 0.22 & 0.25 & 0.29 & 0.14\\

    % \OURSSgnOnly & 0.75 & 0.84 & 0.84 & 0.64 & 0.53 & 0.70 & 0.90 & 0.23 & 0.62 & 0.16 & 0.10 & 0.58 & 0.77 & 1.11 & 0.70 & 0.84 & 0.55 & 0.82 & 0.18 & 0.14 & 0.90 & 0.38\\
    %  \OURSMagOnly & 0.82 & 0.71 & 1.10 & 0.63 & 0.34 & 0.82 & 0.77 & 0.27 & 0.51 & 0.20 & 0.11 & 0.44 & 0.88 & 0.91 & 0.52 & 0.92 & 0.46 & 0.53 & 0.20 & 0.13 & 1.09 & 0.57\\
    % \OURS & 0.90 & 0.95 & 1.27 & 0.94 & 0.53 & 0.84 & 0.90 & 0.31 & 0.62 & 0.20 & 0.14 & 0.60 & 1.04 & 1.15 & 0.75 & 1.19 & 0.72 & 0.88 & 0.27 & 0.16 & 1.36 & 0.60\\
    % \midrule
    
    \multicolumn{23}{c}{Ratio of unique states visited $\uparrow$} \\
    
    \bottomrule
    \end{tabular}
}
\vspace{-5mm}
\end{table*}

\begin{figure}[t] 
    \vspace{2mm}
    \centering
    \includegraphics[width=0.98\linewidth]{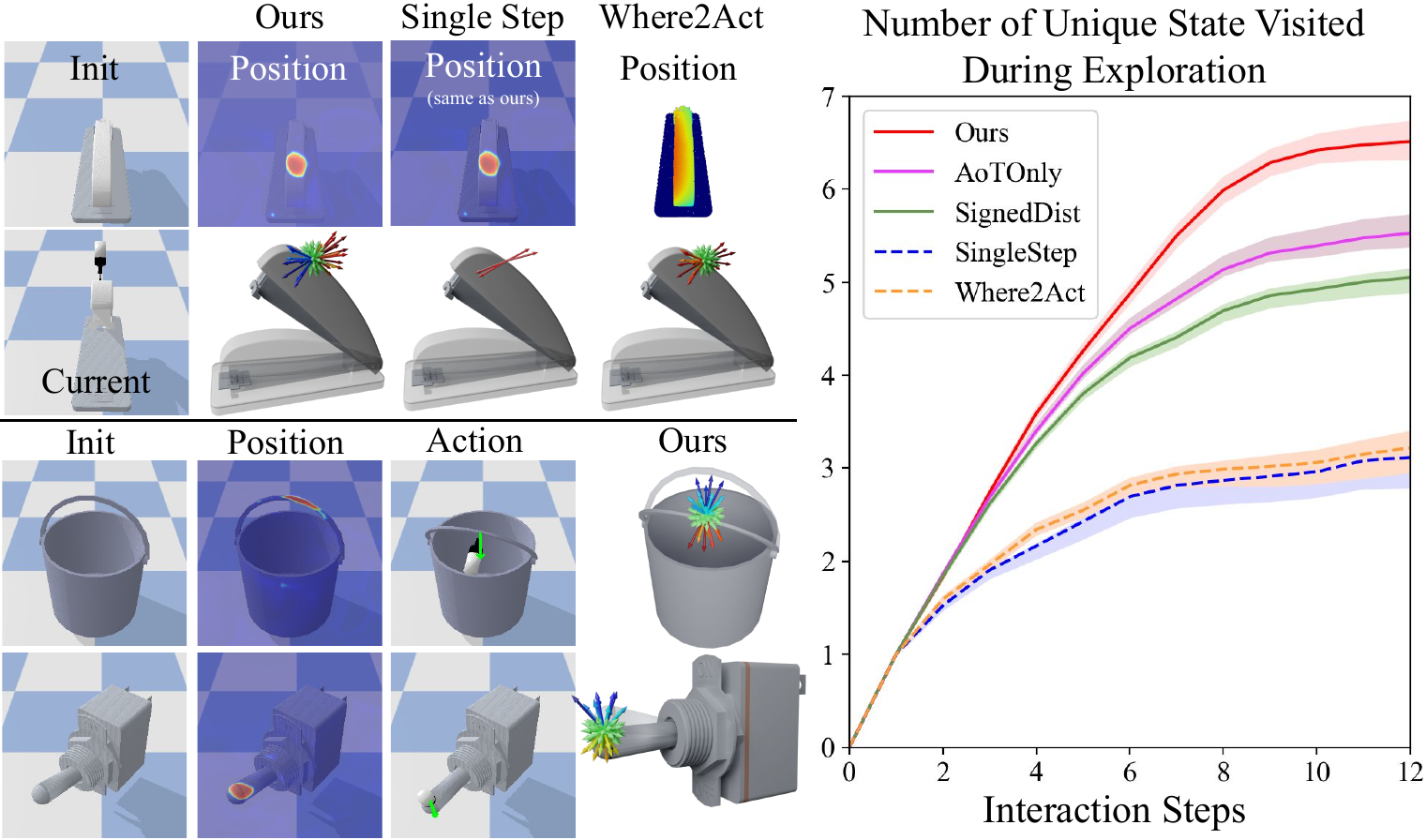}
    \caption{\textbf{Open-ended state exploration.} Arrow length indicates the inferred distance value, color indicates the inferred AoT label. We visualized the uniform samples to better illustrate the AoT distribution. 
    \textbf{(Left)} Qualitative comparisons. All methods are able to choose a suitable position, however, both \OURSSingleStep and \WHERETOACT cannot distinguish between actions that are moving away from or back to initial state (all directions are red) leading to inefficient exploration. In contrast, \OURS is able to infer the correct AoT labels, hence, select the correct action to explore novel states. \textbf{(Right)} Number of unique state visited up to each step using different exploration strategy (laptop testing instances). The error bar is measure with five random seeds. } %The category is laptop (testing instances) and each model is tested five times with different random seeds.
    \label{fig:exploration}
    % https://docs.google.com/drawings/d/12-8np34DY2--hGqXB5jsbR9JEJvpdjloE1dzRr6ayGg/edit?usp=sharing
    \vspace{-6mm}
\end{figure}

\vspace{-1mm}
\subsection{Goal conditioned manipulation with reversed AoT}
\vspace{-0.5mm}
\label{sec:manipulation}

% why goal conditioned 
% how we do it
% key idea swap history with goal and select the negative action 

While open-ended interaction is useful for exploring and collecting information about the environment, most manipulation tasks are goal conditioned -- the policy needs to generate actions that would lead toward a given goal state instead of a random novel state.  
Although the policy is trained with only open-ended exploration, the learned policy can be directly applied to perform goal conditioned manipulation without additional training. 

The key idea for performing the goal-conditioned task is to swap out the initial observation with the goal state observation as the input to the policy. Then by executing the  actions with a reversed Arrow-of-Time (i.e., negative AoT), the policy tries to move object back to the ``past'', which will effectively move the objects toward the goal. 
If the AoT prediction of all direction candidates are non-negative (no blue arrows in Fig. \ref{fig:goal-method}), the trajectory will terminate. 

Apart from choosing the right action direction, another unique challenge for goal-conditioned manipulation is how to choose the correct link to interact when there are multiple movable links on the object (e.g., fridge with double doors in Fig. \ref{fig:goal-method}). 
While the position heatmap predicted by the network covers all movable links, only interacting with the right one can lead to the goal. 
Therefore, to choose a proper position, we first compute a difference mask between the initial and target observation. Then, we multiply the raw position heatmap and the mask to get the filtered position affordance (remove the pixels that are not changed). The final position is selected from the filtered heatmap. 
The algorithm for  goal-conditioned manipulation is illustrated in Fig. \ref{fig:goal-method}.

\section{Evaluation}
\label{sec:evaluation}

Our simulation environment uses objects from PartNet-Mobility \cite{xiang2020sapien} and physics engine from Pybullet \cite{coumans2017pybullet}. We use 12 categories for training and 10 categories for testing. There are 504 training object instances, 132 testing object instances from training categories, and 261 object instances in the testing categories. 
We randomly load an articulated object into the simulation for each interaction session with a randomly initialized pose and joint configurations.

%In the first interaction step, the agent chooses a pixel from image observation. The pixel coordinate will then be converted to a 3D coordinate with its depth value and the end-effector will move to this position and attach to the corresponding link through suction. In each following steps, the agent will select a 3D direction the end-effect will move 0.18(m) along this direction.\todo{more details about the end-effector?} \shuran{yes, one sentence on how you implement the suction gripper so that is it realistic  e.g., suction check with surface normal collision check etc}

%We use Pybullet as our simulation environment, the articulated objects used in simulation are from PartNet-Mobility dataset \cite{xiang2020sapien}. We use 12 categories for training and and 10 categories for testing. In total, there are 636 objects in the training categories and 261 objects in the testing ones. We further divide the training split into 504 training instances and 132 testing instances, and only use the training instances from the training categories to train our networks. For each interaction simulation, we first randomly choose an articulated object from training dataset. We initialize the starting pose randomly for each articulated part and place the object on the ground with random position and rotation. 

% \input{tables/exploration_table-rebuttal}

\vspace{-1.5mm}
\subsection{Open-ended state exploration}
\vspace{-0.5mm}

% \begin{figure}[t] 
%     \includegraphics[width=\linewidth]{figures/where2act-failure.pdf}
%     \vspace{-3mm}
%     \caption{\textbf{Failure Case of Where2Act+Heuristic Policy.} The current action is shown in green, and the last action is shown in transparent brown. In step 4, the baseline chooses a sub-optimal action, which doesn't move the knob. The following steps will also be affected by the filtering process, which shows the heuristics' sensitivity to error propagation.}
%     \label{fig:where2act_failure}
%     % https://docs.google.com/drawings/d/1bzwneauB0shMoTVKxNnA6X7cTKxyGT2VbkllXm0wLdA/edit?usp=sharing
%     \vspace{-4mm}
% \end{figure}

We first evaluate \OURS's effectiveness in exploring novel states of an articulated object. 
Being able to effectively explore the possible states of an object without a specific goal is a critical first step for many robot learning algorithms since it is often used to collect the initial observation about the environment to initiate the training. 
While random explorations can be used for simple environments,  they are often not sufficient for tasks involving high-dimensional action space, where the majority of the actions will not change the object joint state in a meaningful way. 

Instead, an \textit{effective} state exploration policy should be able to choose actions that can (1) significantly change the joint state of an object and (2) lead to novel states that have not been visited before. The first property requires the system to understand the object structure, and the second property requires the system to be aware of the interaction history.

\mypara{Metrics.} We use two metrics to evaluate the effectiveness of state exploration:
(1) \ul{Single action effects} --  measures the joint state difference before and after each interaction step {\small $ D = ||\vec{j}_t - \vec{j}_{t-1}||/ \delta $}. 
The threshold of significant state change $\delta $ is 0.15m for prismatic joint and 8.6\degree for revolute joint. 
 %The unit is radian for revolute joint and meter for prismatic joint. 
This metric evaluates whether the algorithm can choose the action that would change the state of the object most significantly.  
(2) \ul{Novel state visited} -- measures the ratio between the number of unique states visited among all interaction steps: $\mathrm{ratio} = {\mathrm{\# unique\_states}}/{\mathrm{\# steps}}$. Two states consider the ``same'' when the object's joint difference is less than $\delta$. 
This metric evaluates whether the algorithm is aware of the interaction history and chooses the action leading to novel states that have not been visited before.
\mypara{Algorithm comparisons.} We compare our final model  with the following alternative approaches: 
\begin{table*}[t]
% \vspace{1.5mm}
\caption{Goal conditioned manipulation$^1$
\label{tab:manipulation}}
\centering
{\scriptsize %\color{blue}
    \setlength\tabcolsep{2pt}
    \begin{tabular}{l|cccccccccccc|cccccccccc}
    \toprule
      & \multicolumn{12}{c|}{Novel Instances in Train Categories} &  \multicolumn{10}{c}{Test Categories} \vspace{0.5mm} \\
    % \midrule
     & 
    \includegraphics[width = 0.032\linewidth]{icon/noun_Fridge_1875643.png} &
    \includegraphics[width = 0.032\linewidth]{icon/noun_folding_chair_2151184.png} & 
    \includegraphics[width = 0.032\linewidth]{icon/noun_Laptop_2291662.png} &
    \includegraphics[width = 0.032\linewidth]{icon/noun_Stapler_2557851.png} &
    \includegraphics[width = 0.032\linewidth]{icon/noun_trashcan_2244926.png} &
    \includegraphics[width = 0.032\linewidth]{icon/noun_Microwave_1041630.png} &
    \includegraphics[width = 0.032\linewidth]{icon/noun_Toilet_3121.png} &
    \includegraphics[width = 0.032\linewidth]{icon/noun_window_3203560.png} &
    \includegraphics[width = 0.032\linewidth]{icon/noun_Cabinet_2881254.png} &
    \includegraphics[width = 0.032\linewidth]{icon/noun_switch_3674178.png} &
    \includegraphics[width = 0.032\linewidth]{icon/noun_Kettle_3002541.png} &
    \includegraphics[width = 0.032\linewidth]{icon/noun_Graph_281095.png} &
    \includegraphics[width = 0.032\linewidth]{icon/noun_Box_1650724.png} &
    \includegraphics[width = 0.032\linewidth]{icon/noun_flip_phone_143303.png} &
    \includegraphics[width = 0.032\linewidth]{icon/noun_dish_washer_3307528.png} &
    \includegraphics[width = 0.032\linewidth]{icon/noun_safe_1202915.png} &
    \includegraphics[width = 0.032\linewidth]{icon/noun_Oven_7255.png} &
    \includegraphics[width = 0.032\linewidth]{icon/noun_Laundry_1976992.png} &
    \includegraphics[width = 0.032\linewidth]{icon/noun_Table_59987.png} &
    \includegraphics[width = 0.032\linewidth]{icon/noun_kitchen_pot_3363643.png} &
    \includegraphics[width = 0.032\linewidth]{icon/noun_bucket_3910054.png} &
    \includegraphics[width = 0.032\linewidth]{icon/noun_Door_1549119.png}
    \vspace{0.5mm} \\
    
    %\midrule
    \INVERSE \cite{agrawal2016learning} & 0.30 & 0.21 & 0.32 & 0.31 & 0.27 & 0.17 & 0.28 & \textbf{0.09} & \textbf{0.27} & 0.25 & \textbf{0.09} & 0.34 & \textbf{0.25} & 0.32 & 0.09 & 0.17 & 0.27 & \textbf{0.15} & \textbf{0.21} & \textbf{0.00} & 0.51 & 0.27\\
    
    \OURSSgnOnly & 0.23 & \textbf{0.18} & 0.12 & 0.22 & 0.32 & 0.18 & 0.15 & 0.16 & 0.32 & 0.38 & 0.12 & 0.08 & 0.30 & 0.05 & 0.07 & 0.18 & 0.31 & 0.18 & 0.27 & \textbf{0.00} & 0.31 & 0.18\\
    
    \OURSMagOnly & 0.26 & 0.24 & 0.11 & 0.20 & 0.35 & 0.19 & 0.22 & 0.15 & 0.41 & 0.44 & 0.13 & 0.12 & 0.32 & 0.09 & 0.11 & 0.20 & 0.34 & 0.22 & 0.31 & \textbf{0.00} & 0.30 & 0.22 \\
    
    \OURS & \textbf{0.20} & 0.19 & \textbf{0.05} & \textbf{0.19} & \textbf{0.23} & \textbf{0.16} & \textbf{0.12} & 0.13 & 0.28 & \textbf{0.21} & 0.11 & \textbf{0.04} & 0.26 & \textbf{0.03} & \textbf{0.06} & \textbf{0.15} & \textbf{0.21} & 0.16 & 0.22 & \textbf{0.00} & \textbf{0.22} & \textbf{0.17}\\

    \midrule
    \multicolumn{23}{c}{Normalized distance to target $\downarrow$} \\
    \midrule
    
   \INVERSE \cite{agrawal2016learning} & 0.43 & 0.68 & 0.72 & 0.55 & 0.63 & \textbf{0.89} & 0.78 & \textbf{0.65} & 0.61 & 0.52 & \textbf{0.83} & 0.54 & 0.67 & 0.59 & 0.80 & 0.73 & 0.58 & \textbf{0.83} & 0.67 & \textbf{1.00} & 0.39 & 0.68 \\
    \OURSSgnOnly & 0.46 & 0.76 & 0.81 & 0.71 & 0.52 & 0.83 & 0.86 & 0.52 & 0.45 & 0.43 & 0.81 & 0.88 & 0.61 & 0.86 & \textbf{0.86} & 0.7 & 0.52 & 0.77 & 0.6 & \textbf{1.00} & 0.50 & 0.77 \\
    \OURSMagOnly & 0.47 & 0.59 & 0.84 & \textbf{0.75} & 0.48 & 0.88 & 0.75 & 0.52 & 0.49 & 0.37 & 0.78 & 0.83 & 0.58 & 0.84 & 0.83 & 0.69 & 0.46 & 0.71 & 0.57 & \textbf{1.00} & 0.52 & 0.74 \\
    \OURS & \textbf{0.67} & \textbf{0.78} & \textbf{0.90} & 0.73 & \textbf{0.68} & 0.86 & \textbf{0.90} & 0.58 & \textbf{0.63} & \textbf{0.57} & 0.79 & \textbf{0.94} & \textbf{0.68} & \textbf{0.89} & \textbf{0.86} & \textbf{0.76} & \textbf{0.62} & 0.80 & \textbf{0.68} & \textbf{1.00} & \textbf{0.57} & \textbf{0.79} \\
    \midrule
    \multicolumn{23}{c}{\update{Success rate $\uparrow$}} \\
    
    % \midrule
    
    % \OURS* & 0.22 & 0.19 & 0.06 & 0.22 & 0.24 & 0.19 & 0.15 & 0.14 & 0.29 & 0.24 & 0.11 & 0.07 & 0.26 & 0.05 & 0.07 & 0.15 & 0.22 & 0.16 & 0.23 & 0.00 & 0.25 & 0.19 \\
    \bottomrule
    \end{tabular}
}
\vspace{-3mm}
\end{table*}

\footnotetext{Categories: fridge, folding chair, laptop, stapler, trashcan, microwave, toilet, window, cabinet, switch, kettle, toy, box, phone, dish washer, safe, oven, washing machine, table, kitchen pot, bucket, door.}
\begin{figure*}[t]
    \centering
    \includegraphics[width=\linewidth]{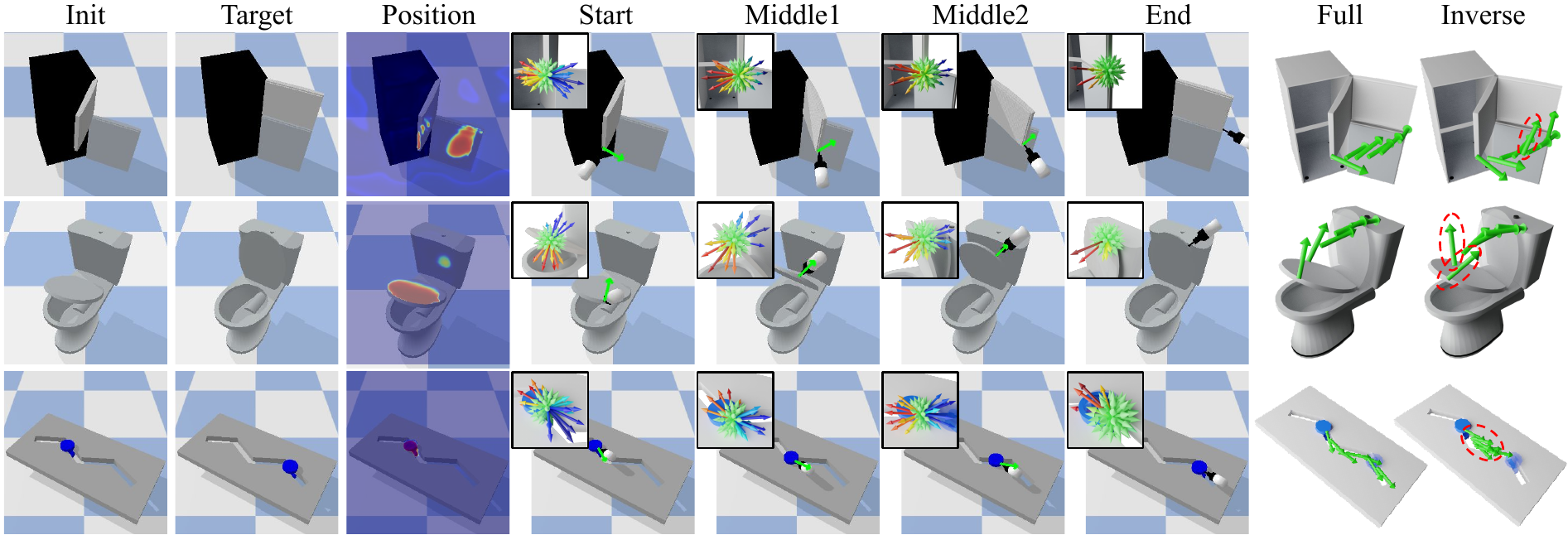}\vspace{-1mm}
    \caption{\textbf{Goal conditioned manipulation results.} 
    At the beginning or in the middle of a trajectory,  the action candidates have positive (red) and negative (blue) AoT labels. To move toward the goal, the policy selects the action with the largest distance prediction and a negative AoT label (the longest blue arrow) to execute. When reaching the goal state (current and goal state are similar), the AoT labels turn non-negative for all actions since all actions will either make no change or move further away from the goal state. 
    The [Inverse] model (right-most column) often chooses sub-optimal action directions (highlighted by red dash circles) at the beginning of the interaction sequence where the current observation is far away from the goal states.}
    \label{fig:manipulation} \vspace{-5mm}
    % https://docs.google.com/drawings/d/1CNBE6hht9rhv4DJuuS46Jm85UsdbsblOfAjzsmRmGSQ/edit?usp=sharing
\end{figure*}

\noindent\textbullet~ \WHERETOACT \cite{mo2021where2act}: This algorithm takes the current observation as input and selects single-step action. The model is with binary-classification loss where the action is positive if only the moving distance is larger than a threshold.\\
\textbullet~ \WHERETOACTHP:  an additional heuristic that filters out actions that has a larger than 90\degree angle with last-step action. This heuristic helps to avoid back-and-forth actions, however cannot be applied for goal-conditioned manipulation.  \\
\textbullet~ \OURSSingleStep: Single-step version of our method that only takes the current observation as input. \\
\textbullet~ \OURSSgnOnly: This method only outputs AoT label for each action without the distance inference.\\
\textbullet~ \OURSMagOnly: Instead of inference AoT and distance as separate outputs, this method infers signed distance by multiplying the AoT and distance value $r_\textrm{singed} =r_\textrm{AoT} \cdot r_\textrm{dist}$

\mypara{Results and analysis.}
Quantities and qualitative results are summarised in  Tab. \ref{tab:exploration} and Fig. \ref{fig:exploration}. 

\ul{Effect of the AoT prediction.} Both [ \WHERETOACT] and [ \OURSSingleStep] only take the current observation as input and infer actions for one step; hence, they do not need to understand the interaction history. 
From Tab. \ref{tab:exploration} we can see that [ \WHERETOACT] is able to achieve similar performance in ``single action effects'', however, both [ \WHERETOACT] and [ \OURSSingleStep] cannot effectively explore novel states with more interaction steps. Since both algorithms are not aware of interaction history,  we observe that the policy often selects actions that would manipulate the object link back-and-forth instead of exploring new possible object states. 
When combined with the heuristic the algorithm [ \WHERETOACTHP]  can avoid back-and-forth action, however, it is sensitive to error propagation, where one sub-optimal action would affect all following steps through the filtering process, results in worse performance. 
Fig. \ref{fig:exploration} shows examples of action prediction results for [ \OURS]. With just the Arrow-of-Time prediction, [ \OURS] is able to identify the actions that would always move the object from the past states (i.e., red arrows); therefore, it is able to visit novel states much more frequently. When combined with heuristic filter, the performance improves slightly. 

\ul{Effect of the distance prediction.} Compared to [ \OURSSgnOnly], we can observe that by explicitly predicting the distance value for each action candidate, [ \OURS]  can better differentiate between different action directions and choose the optimal action direction that would introduce larger state changes.  As a result,  [ \OURS ] can achieve a better ``single action effect'' for all object categories, leading to more efficient state exploration when considering the entire sequence. 

\ul{Effect of decomposing AoT and distance prediction.} Different from  [ \OURSMagOnly] that directly predicts a signed distance value that combines the AoT and distance, [ \OURS ]  decompose its output as an AoT label (trained with classification) and a distance value (trained with regression). 
This decomposition helps the algorithm better disentangle these two concepts, allowing the algorithm to achieve more accurate predictions for both.  As a result,  [ \OURS ] can achieve better performance in both metrics.

\vspace{-1.5mm}
\subsection{Goal conditioned manipulation}
\vspace{-0.5mm}

In this experiment, we evaluate \OURS's performance in the task of goal-conditioned manipulation. 
Given a target state in the form of an RGB-D image, the task is to infer a sequence of actions that manipulate the object toward the target state and halts when the object reaches the target state. 

\mypara{Metrics.} The performance for this task is measured by \update{(1)} normalized distance $\mathcal{E}_\mathrm{goal}$ to target state after interaction: {\small $\mathcal{E}_\mathrm{goal} = ||\vec{j}_\mathrm{end} - \vec{j}_\mathrm{goal}|| / ||\vec{j}_\mathrm{goal} - \vec{j}_\mathrm{init}||$}, where $\vec{j}$ is vector of object's joint state. \update{(2) success rate, where a successful case is defined as the normalized distance to the goal state is smaller than 0.1.}
To make the task more challenging, the initial and goal states are selected from the upper and lower limits of the joint. The initial state may be moved to ensure the task can be accomplished in 15 steps.

\mypara{Algorithm comparisons.} We compare with the [ \INVERSE] model proposed by Agrawal et al. \cite{agrawal2016learning}, a single-step inverse model for goal-conditioned manipulation. Each step takes the current and goal observation as input and predicts the action that would change the state from the current state to the goal state. This model is trained on the same state-action pairs ($s_t, s_{t+1}, a_t$) as our method, and the action output is trained with direct regression loss.

\begin{figure}
    \vspace{2.5mm}
    \begin{center}
    \includegraphics[width=\linewidth]{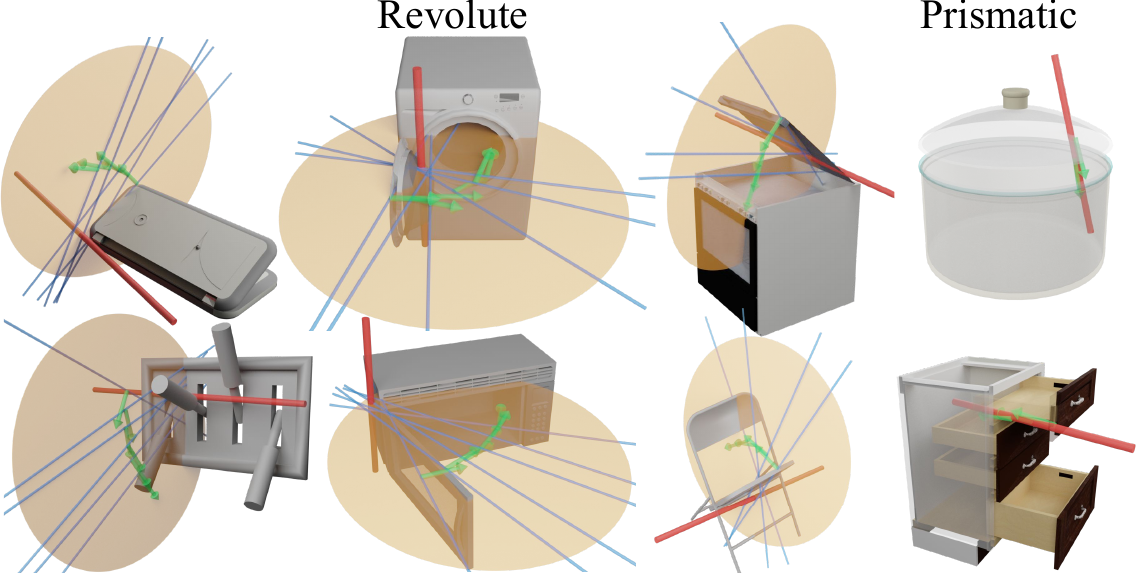}
    \vspace{-4mm}
    \end{center}
    \caption{\textbf{Action $\rightarrow$ Articulation.} The joint axes (red) are inferred from the actions selected by the learned policy (green), which indicates the system's implicit understanding about the objects' articulation structure. }
    \label{fig:structure}
    %https://docs.google.com/drawings/d/1KFK0wOr09HHy2aT2wxEs8VyiYhtOL_p6z4_jfwkgLTI/edit?usp=sharing
    \vspace{-6mm}
\end{figure}
\mypara{Results and analysis.}
Tab. \ref{tab:manipulation} shows that comparing to prior works [ \INVERSE]  and other alternative approaches, [ \OURS] is able to achieve more precise goal-conditioned manipulations by moving the object to a state that is closer to the target (lower {\small $\mathcal{E}_\mathrm{goal}$} value). 
From the qualitative comparisons in Fig. \ref{fig:manipulation} we can observe that the performance of the [ \INVERSE] model is much worse at the beginning of the interaction, where the algorithm often selects sub-optimal action directions that make less progress towards the goal (actions highlighted in red dash circle). Since the [ \INVERSE] model only takes consecutive observations as input during training, it struggles to handle long-horizon manipulation tasks, where the current observation is far away from the goal states.  
Similar to exploration experiments, we observe that [ \OURSSgnOnly] often chooses sub-optimal action direction as it is unaware of the actual magnitudes (i.e., distance) of different action effects. 
%
% Furthermore, the small gap between [\OURS ] and [ \OURS* ] shows that the AoT label can distinguish whether the object has reached the target state and terminate the manipulation trajectory properly.

\vspace{-0mm}
\subsection{Inferring objects' articulation structure from interactions}
\vspace{-0.5mm}

We hypothesize that one of the requirement for learning a universal policy for \textit{any} articulated object is the ability to understand the object's underlying articulation structure and how this structure react to different actions. 
Hence, the action selected by the policy should also, in return, reflects its belief on the objects' structure. 
For example, we often apply forces along the axis for prismatic joints while applying actions perpendicular to the rotation axis for revolute joints.

To visualize the policy's implicit belief about the object's structure,  we compute the joint parameters inferred from the actions selected by the policy.
To compute the prismatic joint, we simply take the average of the action directions. 
To compute the revolute joint, we first compute a common action plane in the 3D space (brown plane in Fig \ref{fig:structure}). The normal direction of the plane $\vec{n} \in R^3$ is chosen as $\min_{||\vec{n}||=1} \sum_{t=1}^{T} |\vec{n} \cdot a_t|$, where $a_t$ is the action direction in each interaction step.  
Then we vote for the axis position by computing the interaction sections between the directions perpendicular to all the actions in the common plane (blue lines in Fig \ref{fig:structure}). Finally, the final axis position is voted among the intersection points between each pair of the perpendicular lines.
Fig. \ref{fig:structure} shows examples of inferred joint parameters for objects with different articulation structures (red lines).

We also quantitatively evaluate the inferred joint parameters. 
While the algorithm has never been supervised on any of the joint parameters, it is able to estimate the joint axis orientation with an average error $< 11.6$\degree~for revolute joints and  $<32.2$\degree~ for prismatic joints.  Note that the error in prismatic joint estimation is higher since these objects often has higher tolerance on the sub-optimal action directions.

\vspace{-1.5mm}
\subsection{Real-world experiment}
\vspace{-0.5mm}

\begin{figure*}[t]
    \vspace{1mm}
    \centering
    \includegraphics[width=\linewidth]{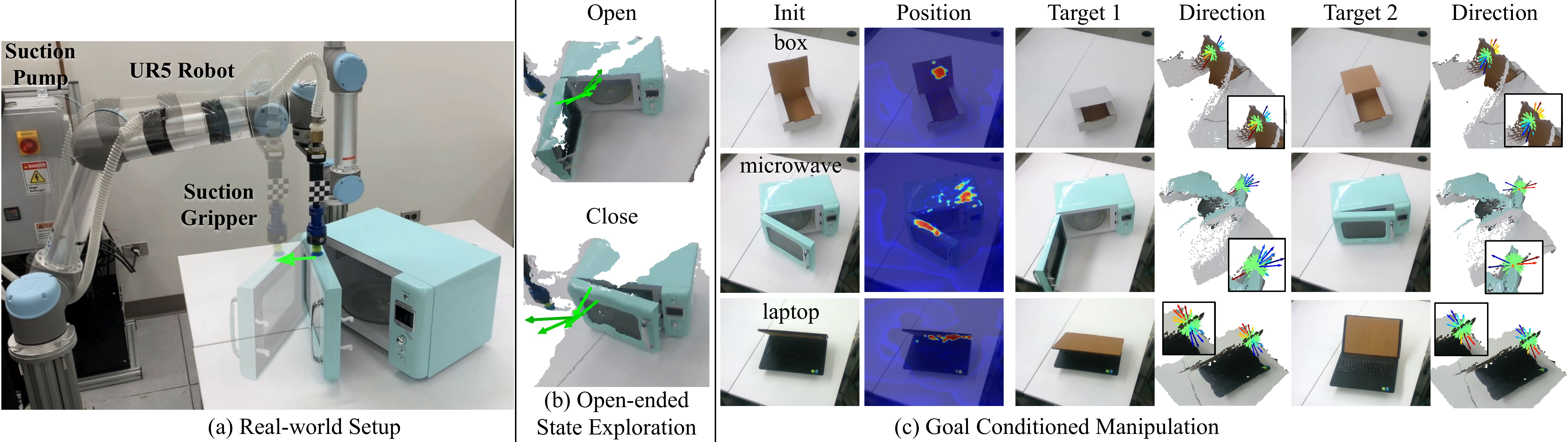}\vspace{-1mm}
    \caption{\textbf{Real-world experiment.} We test the model trained in simulation on a real-world platform. \textbf{(a)} We an RGB-D camera to capture visual observation and a UR5 with a suction gripper  for manipulation. \textbf{(b)} Action trajectory. \textbf{(c)} For each object, we visualize the inferred action position and direction for two different target states. To move toward the goal, the policy will select the action with the largest distance prediction and a negative AoT label (the longest blue arrow) to execute.}
    \label{fig:real-result} 
    %https://docs.google.com/drawings/d/1JvzE4kKjRDTQnkWRQjTo8FVs6ihPRh-RYw1QRT79er0/edit?usp=sharing
    \vspace{-5mm}
\end{figure*}

Finally, we validate our method on a real-world platform with a calibrated RGB-D camera (Intel RealSense D415), a UR5 robot, and a suction gripper. Fig. \ref{fig:real-result} (a) shows the real-world setup. In this experiment, we directly tested \OURS trained in simulation on four different objects -- box, laptop, microwave, and stapler. The inferred action trajectories to open and close the microwave are shown in Fig. \ref{fig:real-result} (b). The qualitative result of goal-conditioned manipulation shown in Fig. \ref{fig:real-result} (c) demonstrates that the trained model is able to infer proper grasping positions and action directions for different objects and goal conditions.  \update{While performing large-scale real-world training for \OURS can still be challenging, we believe these results demonstrate the promises of the proposed method in real-world applications. We observed that there are a few real2sim gaps that could impact real-world performance. For example, the noise captured by the depth camera could affect direction inference. For objects don't have a fixed base (e.g., microwave), they might experience unexpected movements during interactions, and therefore negatively impact the algorithm performance.} \finalupdate{In addition, our policy doesn't consider real robot situation, for example, whether the grasping position can be reached by a real robot, the moving trajectory is safe, the grasping surface is flat enough for a robust suction. All these issues about real robot platforms should be considered in our policy in future works.}
\begin{figure}[t]
    \centering
    \vspace{2.5mm}
    \includegraphics[width=\linewidth]{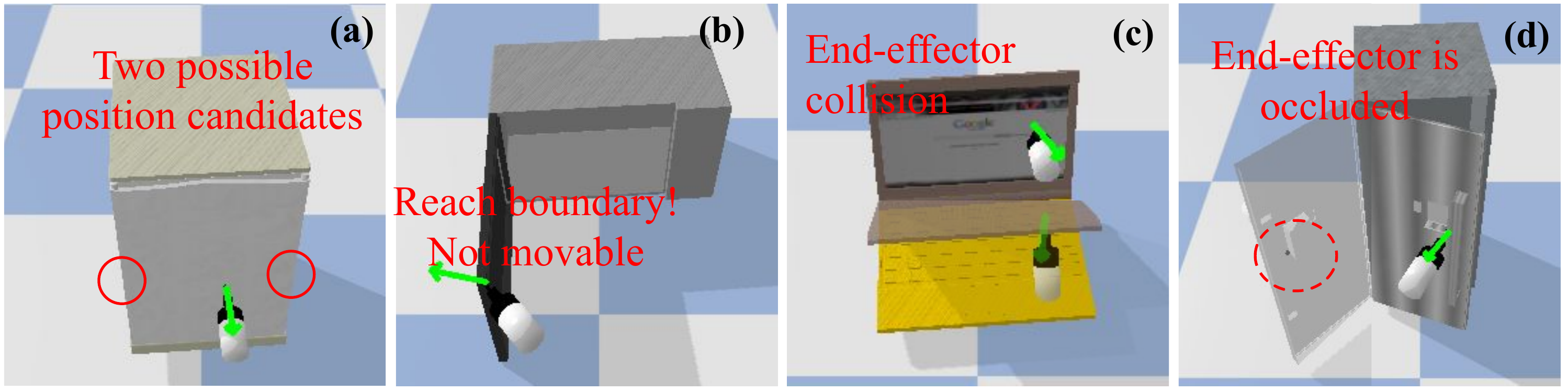}
    \caption{\textbf{Typical failure cases.}}
    \label{fig:failure}
    % https://docs.google.com/drawings/d/1aALJQ57m5YkBbfySDuyIR15_gnXH6C5MV0qHFtRTs9c/edit?usp=sharing
    \vspace{-6mm}
\end{figure}

\vspace{-3mm}\update{\subsection{Limitations and failure cases}}
\label{sec:failure}
\update{
\textbf{Assumptions:} To allow goal-conditioned manipulation with reversed AoT actions, we assume the action trajectories are bi-directional in time (i.e., they are valid in either direction). While this assumption is true for most articulated objects, it does not apply to irreversible actions such as locking. %Furthermore, goal-conditioned manipulation will fail if this is the first contact with the object, because the goal state can only be input as an image with the same scene. \shuran{I don't understand, what is this??}
In addition, our system assumes the agent uses a suction-based end-effector, which can provide robust grasps for a large variety of objects and is widely used in many real-world robotics systems. However, the policy cannot generalize to other grippers that requires more precise grasp poses. %For example, while the suction gripper can grasp the door frame at almost any position, parallel jaw grippers can only grasp the door from its handle. 
Finally, our system assumes there is only a single articulated object with 1 DoF  prismatic or revolute joint on a planar surface, and the goal state can only be input as an image with the same scene.
}

\update{
\textbf{Failure Cases:} Fig.~\ref{fig:failure} case (a) is ambiguous in position selection since the door could be opened from both sides, where the policy chooses to drag the middle of the door. In case (b), the selected action can't change the object state since the microwave's door reaches boundary. However, the joint range can't be easily inferred from observation since some microwaves can be opened up to 180\degree. In case(c), policy infers actions that will cause collisions between the end-effector and the object. In case(d), the end-effector is occluded after interactions. While a human is able to change the viewpoint for better observation, our agent uses a fixed camera position and therefore not robust for occlusion. Both (c) and (d) could be addressed by better modeling the agent's embodiment including end-effector and camera placement.  
}

\vspace{2mm}
\section{Conclusion}
We introduce the Universal Manipulation Policy Network (\OURS) -- a single image-based policy network that infers closed-loop action sequence for manipulating articulated objects. 
The policy is trained with self-guided exploration without human demonstrations, scripted policy, or pre-defined goal conditions. 
Our experiment results demonstrate that the learned policy is able to perform well in both open-ended exploration and goal-conditioned manipulation and outperforms alternative approaches in both tasks.

\bibliographystyle{IEEEtran}
\bibliography{references}

\newpage
\renewcommand{\thesection}{A.\arabic{section}}
\renewcommand{\thefigure}{A\arabic{figure}}
\renewcommand{\thetable}{A\arabic{table}}
\setcounter{section}{0}
\setcounter{figure}{0}
\setcounter{table}{0}

%\appendix
\section*{Appendices}
Please check out our supplementary video for additional results and comparisons. In this document we present additional details on the object dataset \S \ref{supp:sec:object}, network architectures \S \ref{supp:sec:network}, sampling strategy \S \ref{supp:sec:action_sample}, and quantitative results for each object category. 

\section{Additional details: object dataset} \label{supp:sec:object}
\begin{figure*}[ht] 
    %\vspace{-3mm}
    \includegraphics[width=\linewidth]{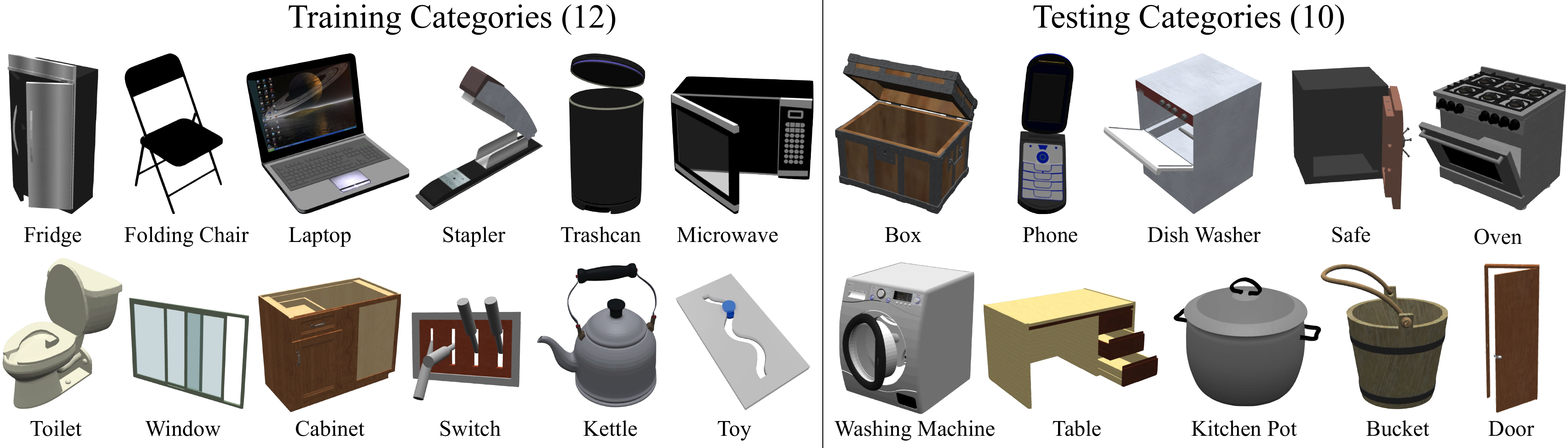}
    %\vspace{-3mm}
    \caption{\textbf{Training and testing categories.}} 
    \label{fig:object}
    % https://docs.google.com/drawings/d/15uPZLzsL3vGDR4qGFH_VXwyUZyICejxPu6J74KT2Ofg/edit?usp=sharing
    \vspace{-0mm}
\end{figure*}

Our experiments 11 training categories and 10 testing categories from PartNet-Mobility dataset \cite{xiang2020sapien}. Most of the objects in PartNet-Mobility have relatively simple joint configurations where the desired action trajectories often have a smooth change of direction.
To test whether our algorithm is able to produce more complex action trajectories, we add an additional object category, ''Toy'' (see Fig. \ref{fig:object}), which requires the policy to constantly change output action directions at every interaction steps. 
Within this toy category, we create two types of object instances using Blender, where the wave-shaped instance is used for training, and the zig-zag-shaped instance is used for testing.

Detailed instance statistics and their joint types for each object category are listed in Tab.~\ref{tab:object}. Note that many object instances contain both revolute and prismatic joints. 
For each category, the maximum number of interaction steps $N_\textrm{step}$ is determined by the average joint range divided by $\delta$. Fig.~\ref{fig:object} presents example instances from each object category.

\section{Additional details: network structure} \label{supp:sec:network}
\textbf{PositionNet.}
Given a visual observation captured by an RGB-D camera, we first calculate world coordinates for each pixel using depth value. Surface normals are then estimated via KD Tree searching. Next, RGB-D image, world coordinates, and surface normals are concatenated (size equals 10$\times$480$\times$640) and then fed into PositionNet with a U-Net architecture. The PositionNet applies four down-sample blocks with 32, 64, 128, and 256 channels, followed by four up-sample blocks with 128, 64, 32, and 2 channels. Each down-sample(or up-sample) block includes a max-pooling(or bilinear interpolation) layer and two 3$\times$3 convolution layers with ReLU. Finally, pixel-wise softmax is applied, and the output tensor is position affordance with a size of 2$\times$480$\times$640.

\textbf{DirectionNet.} DistNet takes the current observation $o_t$ as input and applies seven 3$\times$3 convolution layers with 32, 64, 128, 256, 512, 512, and 512 channels. Max pooling is also applied except for the first layer. The output tensor with a size of 512$\times$7$\times$10 is then flattened as an embedding vector $\chi(o_t)$. DistDecoder takes the embedding vector $\psi(o_t)$ and action $\hat{a}^\mathrm{dir}$ as input. A two-layer MLP with both 256 dimensions is applied to the embedding vector $\chi(o_t)$. The action is also encoded via a two-layer MLP with both 128 dimensions. These two vectors are then concatenated and fed into a four-layer MLP with dimensions of 1024, 1024, 1024, and 1. Finally, the network outputs a scalar value as the distance prediction. The network architectures of the AoT branch are similar to those of Dist branch, with only two differences. First, the input channel of AoTNet is doubled since the current observation $o_t$ and initial observation $o_0$ are concatenated and fed into the network. Second, the output dimension of AoTDecoder is three and followed by softmax to perform a three-way classification.

\section{Effect of direction sampling strategy}
\label{supp:sec:action_sample}

\begin{figure}
    \includegraphics[width=\linewidth]{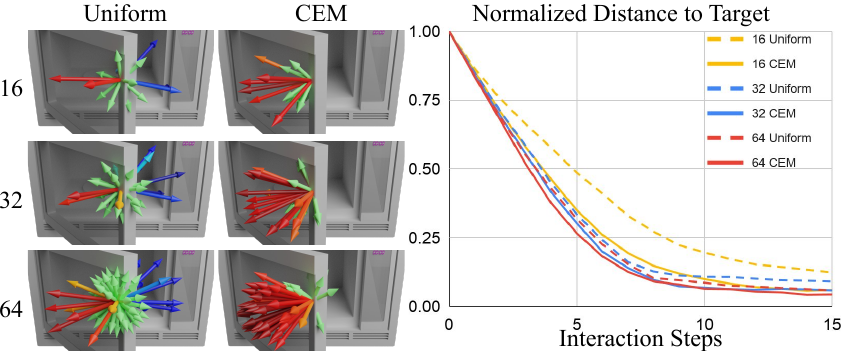}
    \caption{\textbf{Action sample comparison}}
    \label{fig:action_sample}
    % https://docs.google.com/drawings/d/17-87dkYRjEtalF-HnlEyfrX5SajpI5BZfip3SgYiSMg/edit?usp=sharing
    \vspace{-4mm}
\end{figure}

In this experiment, we evaluate the effect of the direction sampling strategy by comparing the Uniform and CEM direction sampling with a different number of samples.  
Fig. \ref{fig:action_sample} (left) visualizes sampled actions from different strategies. Under the same number of samples, CEM can provide denser candidates in the region of interest, making it more likely for the model to select a direction of higher quality.  Fig. \ref{fig:action_sample} (right) shows the algorithm's performance in the goal-conditioned manipulation task using different action samples.     
We observe that the performance improves with more action samples and that CEM sampling consistently achieves a better performance than Uniform sampling under the same number of samples. Therefore, our final model uses CEM sampling with 64 action samples.

\begin{table*}[ht]
    \centering
\vspace{-3mm}
\caption{Statistics of the data splits. 
\label{tab:data}}
{\footnotesize
    \centering
    \setlength\tabcolsep{2pt}
    \begin{tabular}{cccccc|ccccc}
    \toprule
    \multicolumn{6}{c|}{Training Categories} & \multicolumn{5}{c}{Testing Categories}\\
    \midrule
    Name & \# train & \# test & $N_\textrm{step}$ & Revolute & Prismatic & Name & \# test & $N_\textrm{step}$ & Revolute & Prismatic\\
    \midrule
    Fridge & 33 & 9 & 12 & $\times$ & & Box & 13 & 10 & $\times$ & \\
    FoldingChair & 16 & 4 & 8 & $\times$ & & Phone & 3 & 12 & $\times$ & \\
    Laptop & 34 & 9 & 12 & $\times$ & & Dishwasher & 41 & 10 & $\times$ & $\times$ \\
    Stapler & 17 & 5 & 15 & $\times$ & & Safe & 28 & 10 & $\times$ & \\
    TrashCan & 36 & 10 & 9 & $\times$ & $\times$ & Oven & 24 & 9 & $\times$ & \\
    Microwave & 8 & 2 & 8 & $\times$ & & WashingMachine & 17 & 9 & $\times$ & \\
    Toilet & 25 & 7 & 7 & $\times$ & & Table & 77 & 7 & $\times$ & $\times$ \\
    Window & 41 & 11 & 6 & $\times$ & $\times$ & KitchenPot & 25 & 3 & & $\times$ \\
    Cabinet & 266 & 67 & 9 & $\times$ & $\times$ & Bucket & 6 & 13 & $\times$ & \\
    Switch & 8 & 2 & 7 & $\times$ & $\times$ & Door & 27 & 10 & $\times$ & \\
    Kettle & 19 & 5 & 3 & $\times$ & $\times$ & & &\\
    Toy & 1 & 1 & 10 & & $\times$ & &\\
    \midrule
    Total & 504 & 132 & & & & Total & 261 & \\
    \bottomrule
    \end{tabular}
    \label{tab:object}
}
\vspace{0mm}
\end{table*}

\section{Additional results: open-ended state exploration} \label{supp:sec:plot}
\begin{figure*}[ht] 
    %\vspace{-5mm}
    \includegraphics[width=1\linewidth]{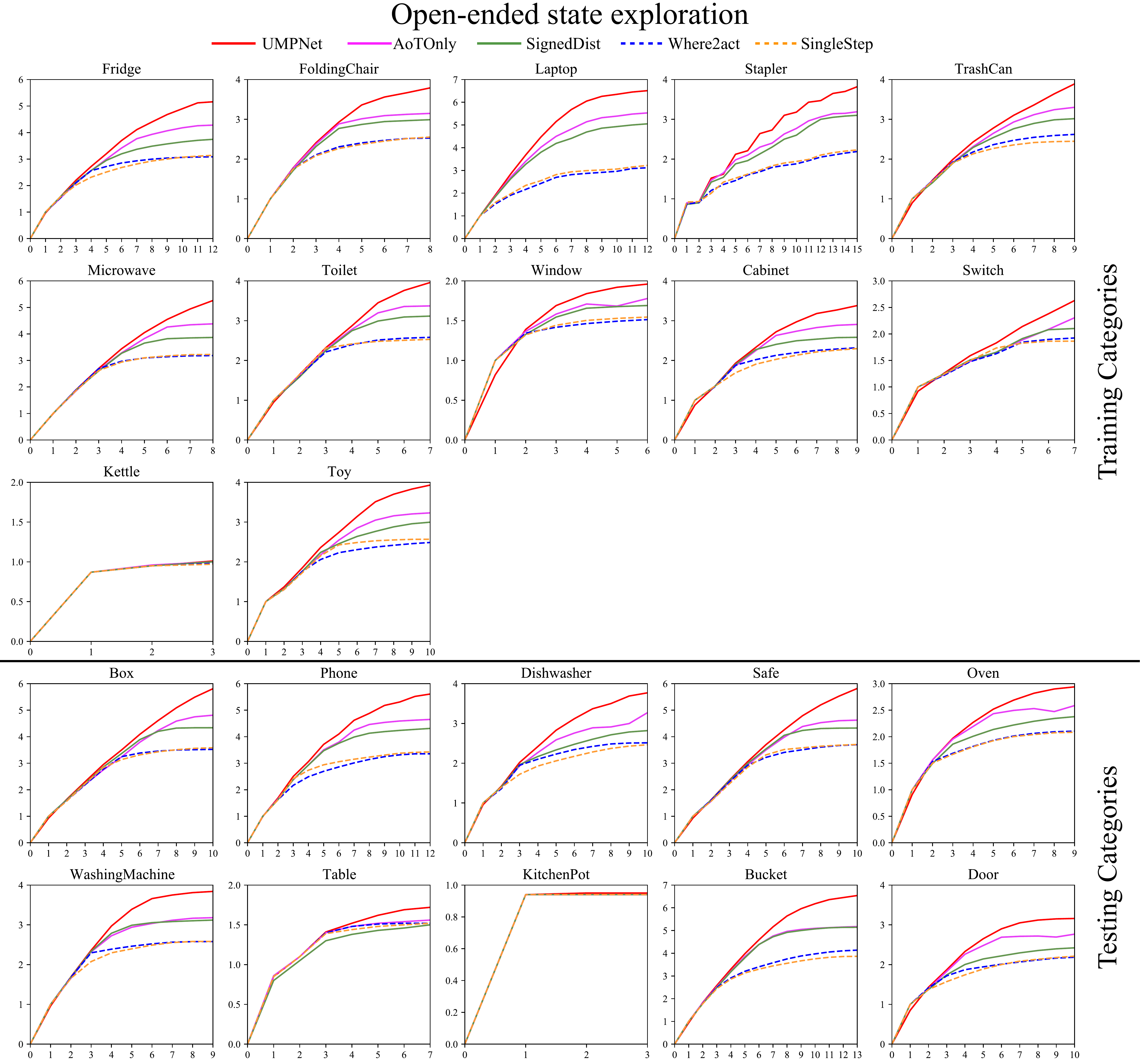}
    %\vspace{-5mm}
    \caption{Number of unique states visited up to each step. UMPNet can achieve significantly better performance when handeling long-sequence interactions since it can choose actions with consistent direction with maximum state change, instead of moving the object link back-and-forth.} 
    \label{fig:exploration_plot}
    %\vspace{-3mm}
\end{figure*}

Fig. \ref{fig:exploration_plot} shows the plot of number of unique states visited during exploration. 

\section{Additional results: goal-conditioned manipulation} \label{supp:sec:plot2}
\begin{figure*}[ht] 
    %\vspace{-5mm}
    \includegraphics[width=\linewidth]{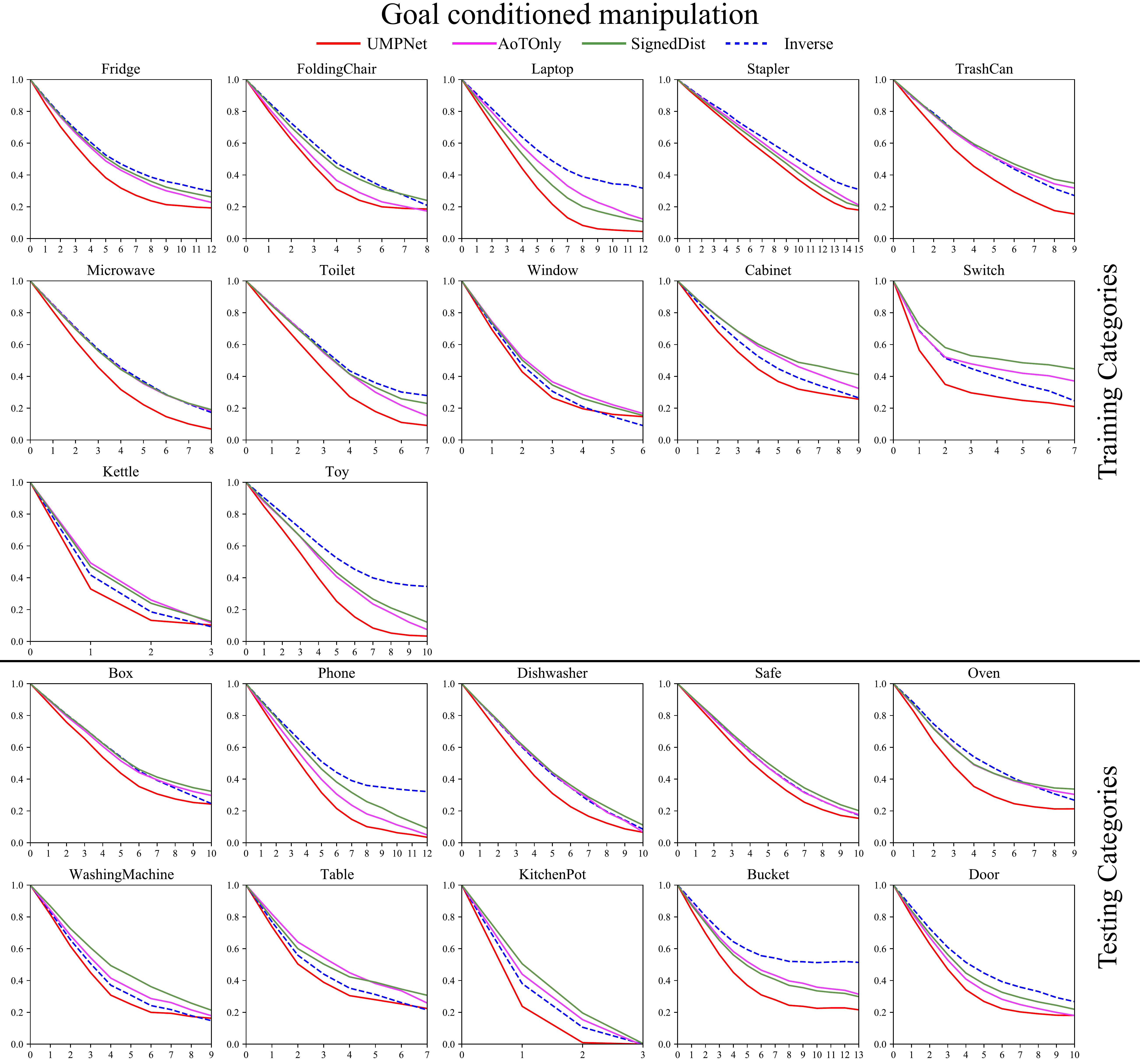}
    %\vspace{-5mm}
    \caption{Normalized distance to target up to each step. UMPNet outperforms other baselines when handling long-horizon goal-conditioned manipulation tasks. } 
    \label{fig:manipulation_plot}
    \vspace{-0mm}
\end{figure*}

Fig. \ref{fig:manipulation_plot} shows the normalized distance to target  in the task of goal-conditioned manipulation. 

\end{document}